\documentclass{article}
\usepackage[table]{xcolor}
\usepackage[final]{corl_2026}
\usepackage{graphicx}
\graphicspath{{figs/}}
\usepackage{subcaption}
\usepackage{wrapfig}
\usepackage{multirow}
\usepackage{booktabs}
\usepackage{makecell}
\usepackage{float}
\usepackage{enumitem}
\usepackage{fontawesome5}
\definecolor{somegray}{rgb}{0.5, 0.5, 0.5}
\newcommand{\darkgrayed}[1]{\textcolor{somegray}{#1}}
\makeatletter
\newcommand*\titleheader[1]{\gdef\@titleheader{#1}}
\AtBeginDocument{%
  \let\st@red@title\@title
  \def\@title{%
    \vskip-3em
    \bgroup\normalfont\large\centering\@titleheader\par\egroup
    \vskip1.5em\st@red@title}
}
\makeatletter
\titleheader{\darkgrayed{This paper has been accepted for publication at the 
\\
Conference on Robot Learning (CoRL), Austin 2026
}
}
\title{
Learning Visuomotor Quadrotor Flight Through Dynamic Narrow Gaps via Differentiable Simulation
}
\author{
  Riccardo Bianco\thanks{These authors contributed equally.
  Correspondence to: \texttt{yunfan@ifi.uzh.ch}, \texttt{rbianco@ethz.ch}.}
  \hspace{0.5cm} Yunfan Ren\footnotemark[1]
  \hspace{0.5cm} Jiaxu Xing
  \hspace{0.5cm} Davide Scaramuzza\\[0.3cm]
  Robotics and Perception Group,
  University of Zurich
}

\begin{document}
\maketitle

\begin{figure}[h]
    \centering
    \vspace{-0.6cm}
    \includegraphics[width=0.95\linewidth]{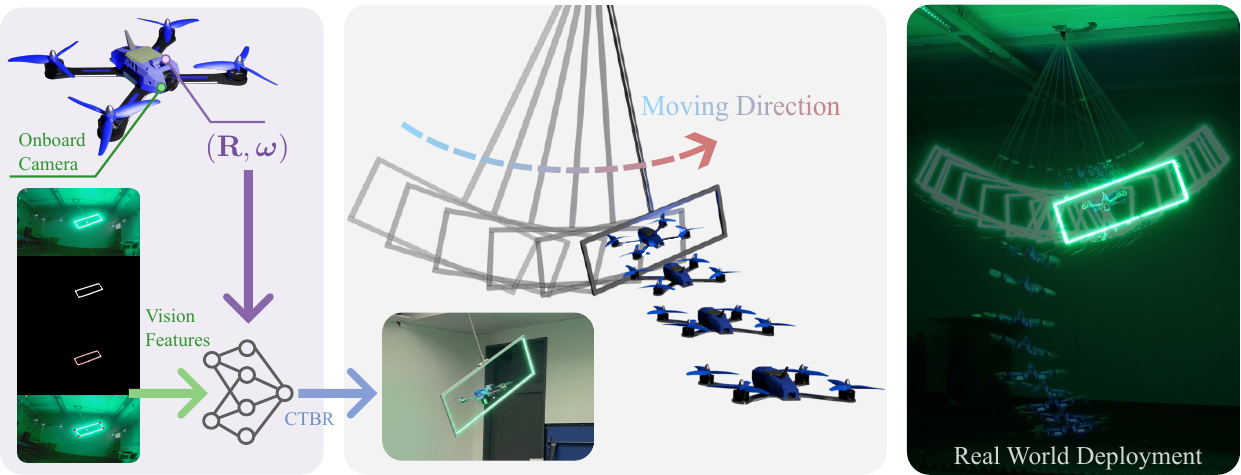}
    \caption{
      Traversal of a narrow \emph{moving} gap using onboard sensing and inertial measurements.
      The policy is trained with differentiable simulation and transferred zero-shot in the real world.
    }
    \vspace{-0.5cm}
    \label{fig:teaser}
\end{figure}

\begin{abstract}
    Threading a quadrotor through a moving narrow gap is a demanding test of agile flight, requiring centimeter-level spatial alignment and precise temporal execution.
    We address this challenge using a control policy that operates directly on onboard visual and inertial measurements to output collective thrust and body-rate commands.
    The task is difficult because success lies on a thin feasible manifold: small errors in timing, position, attitude, or visual observability cause collision.
    Prior planning and learning methods rely on privileged state, teachers, planners, or distillation; learning dynamic narrow-gap traversal directly, from scratch, remains open.
    We introduce a differentiable moving-gap formulation that gates centering, alignment, and smooth collision terms by signed distance to a time-varying gap plane, turning a sparse traversal event into a dense pathwise objective.
    We integrate this formulation into a visual-feature-to-command policy trained in one stage from random initialization, without a teacher, planner, or distillation.
    We validate our approach in both synthetic and real-world experiments.
    In controlled benchmarks, we demonstrate that pathwise gradients significantly outperform score-function gradients for precision-critical tasks, yielding a $99.3\%$ success rate in simulation after only a few minutes of training on a single GPU.
    Furthermore, the policy transfers directly to a real quadrotor, successfully crossing a dynamic narrow gap from varied initial conditions using onboard sensing and inertial measurements, achieving an $80\%$ success rate over 10 physical trials.
    \\
    \faVideo~\textbf{Video:} \url{https://youtu.be/z-txORkhnrg} \\
    \faGithub~\textbf{Code:} \url{https://github.com/uzh-rpg/agile_narrow_gap_diffsim}
\end{abstract}


\section{Introduction}
\label{sec:introduction}

Agile quadrotors can move quickly through confined spaces, but useful motion is often constrained by windows, frames, shelves, machinery, debris, and other structures that leave only narrow passages.
Narrow-gap traversal is therefore a representative primitive for agile flight, requiring the vehicle to coordinate speed with whole-body geometric precision.
Dynamic gaps make the primitive harder: the quadrotor must fit through the aperture and arrive when it is traversable.
We study the onboard visual-inertial version of this problem, learning policies that use vision and inertial measurements to output collective-thrust-and-body-rate (CTBR) commands for static and moving gap traversal.

This setting combines thin spatio-temporal feasibility, partial onboard observations near crossing, and aggressive low-level control.
Prior work has made strong progress on agile quadrotor planning and narrow-gap traversal through model-based trajectory optimization~\citep{falanga2017,liu2018sb,wang2022minco,han2021fastracing,yang2021wholebody,ren2023mr}, learning-based control~\citep{lin2019nn,xie2023learning,xiao2023rl,loquercio2021learning,kaufmann23champion,wu2026sensorimotor}, and differentiable simulation~\citep{wiedemann2023apg,heeg2025visualdiffsim,zhang2025nmi,zhang2026irreg}.
Yet most of these methods rely on static apertures, privileged geometry, or teacher/planner pipelines.
This leaves open whether precision-critical moving-gap traversal can be learned end-to-end from random initialization, without a planner, teacher, or distillation.
A moving aperture changes the problem qualitatively: the policy must decide not only how to cross, but also when to commit to the crossing.

We argue that a central bottleneck is the quality of the learning signal.
Model-free RL methods such as PPO~\citep{schulman2017ppo} estimate score-function (SF) gradients by treating the simulator as a black box, which can be too noisy on thin-manifold tasks.
Pathwise (PW) gradients differentiate through closed-loop rollouts~\citep{kingma2014vae,mohamed2020mcg}, but can become unreliable under stiffness, discontinuities, contacts, or chaotic dynamics~\citep{mora2021pods,suh2022pods,metz2021grads}.
Our key observation is that narrow-gap flight is unforgiving but still smooth: with differentiable quadrotor dynamics and smooth task costs, even failed rollouts indicate how to adjust centering, attitude, and timing.

We build on this observation with a differentiable moving-gap formulation that gates centering, alignment, and smooth collision terms by signed distance to a time-varying gap plane.
Integrated into a visual-feature-to-CTBR policy trained in one stage from random initialization, this formulation turns a sparse traversal event into a dense pathwise objective.
The framework supports privileged-state and onboard visual-inertial observations and requires no demonstrations, trajectory labels, privileged teacher, planner, or distillation.

We validate our approach in simulation and real-world flight.
In simulation, policies train in $0.4$--$5.6$ minutes on a single GPU and reach $\geq\!98.9\%$ success with centimeter-level crossing accuracy.
In matched 1,000-environment, five-seed state-task runs, tuned PPO used $19\times$ more transitions and $45.5\times$ more training time than BPTT on moving gaps when averaging its four converged seeds.
Controlled benchmarks show that PW optimization succeeds with a single rollout per update, whereas SF optimization fails even with one thousand times more samples.
Deployed in the real world, a 10$\times$38\, cm quadrotor traverses a 25$\times$80\, cm moving gap using onboard sensing and inertial measurements, achieving an 80\% success rate over 10 trials.

\textbf{Contributions:}
In summary, the contributions of this work are:
\begin{itemize}[leftmargin=*, nosep, topsep=0pt]
      \item We demonstrate precision-critical vision-based static and moving traversal can be learned end-to-end through differentiable simulation without demonstrations, a teacher, a planner, or distillation.
      \item We release an open-source differentiable-simulation framework for learning observation-to-CTBR policies under both privileged-state and onboard visual-inertial observations, converging in minutes on a single GPU.
      \item We show that pathwise gradients provide useful learning signals on thin-manifold tasks where score-function gradients fail even with orders of magnitude more samples.
      \item We deploy the visual-inertial moving-gap policy on a real quadrotor, closing the loop from differentiable training to physical agile flight.
\end{itemize}

\section{Related Work}
\label{sec:related_work}

\textbf{Narrow-Gap Quadrotor Flight.}
Traversing narrow gaps is a long-standing SE(3)-aware control benchmark, requiring precise attitude commitment under centimeter-level clearance.
Optimization-based methods decompose free space into convex corridors and solve constrained SE(3) trajectory optimizations~\citep{liu2018sb,wang2022minco,han2021fastracing,yang2021wholebody,ren2023mr}; \citet{falanga2017} first demonstrated onboard traversal with active vision but assumed known gap geometry and a hand-crafted estimator.
On the learning side, \citet{lin2019nn} trained an end-to-end planner-controller still requiring known gap pose; \citet{xiao2023rl,chen2023tiv} apply deep RL with curriculum learning and safety-aware exploration but rely on privileged pose information; and \citet{xie2023learning} demonstrate the importance of closing the perception-control loop for static-gap onboard-sensing traversal.
Two concurrent works are closest: \citet{wu2026sensorimotor} bootstrap a
model-free RL policy from a model-based planner and distill it end-to-end, and
\citet{zhang2026irreg} learn from raw depth via differentiable simulation for
irregular static gaps. We differ by (i)~treating \emph{moving} gaps as a
first-class trained task rather than a reactive byproduct; (ii)~learning
end-to-end from random initialization without a teacher, planner, or
distillation, in minutes rather than over ${\sim}1.5$\,h in \citet{wu2026sensorimotor}; and (iii)~isolating why pathwise gradients succeed
where score-function gradients fail.

\textbf{End-to-End Visuomotor Quadrotor Policies.}
Vision-conditioned quadrotor control has advanced through policy distillation and large-scale RL: \citet{loquercio2021learning} fly through cluttered forests by distilling a privileged planner into a vision policy, and \citet{kaufmann23champion} reach champion-level racing via deep RL with high-frequency state estimation. Recent work explores lighter perception interfaces~\citep{xing2024bootstrap, xing2024contrastive}, onboard sensing for angled narrow gaps~\citep{xie2023learning}, and whole-body gap traversal from image masks~\citep{wu2025wholebody}. These policies can match or surpass model-based methods in time-optimal regimes, but typically require expert demonstrations, privileged teachers, or large sample budgets---which our low-variance pathwise gradients from differentiable simulation avoid.

\textbf{Differentiable Simulation for Robot Learning.}
Differentiable simulators optimize control policies directly via analytic gradients. \citet{wiedemann2023apg} showed that analytic policy gradients can train competitive low-level quadrotor controllers, with extensions to vision-conditioned flight~\citep{zhang2025nmi,pan2026learning, ren2026learning, xing2026continual}, locomotion~\citep{song2024quadrupeddiff}, and high-throughput GPU-accelerated frameworks~\citep{diffaero2025,aerialgym2025,visfly2025}. The closest concurrent work, \citet{zhang2026irreg}, applies differentiable simulation with a depth-image renderer to irregular \emph{static} gaps. We target a complementary regime---static \emph{and} dynamic gaps---evaluated against PPO under matched parallelism.

\begin{figure}[t]
    \centering
    \begin{minipage}{0.45\linewidth}
        \centering
        \begin{subfigure}{\linewidth}
            \includegraphics[width=\linewidth]{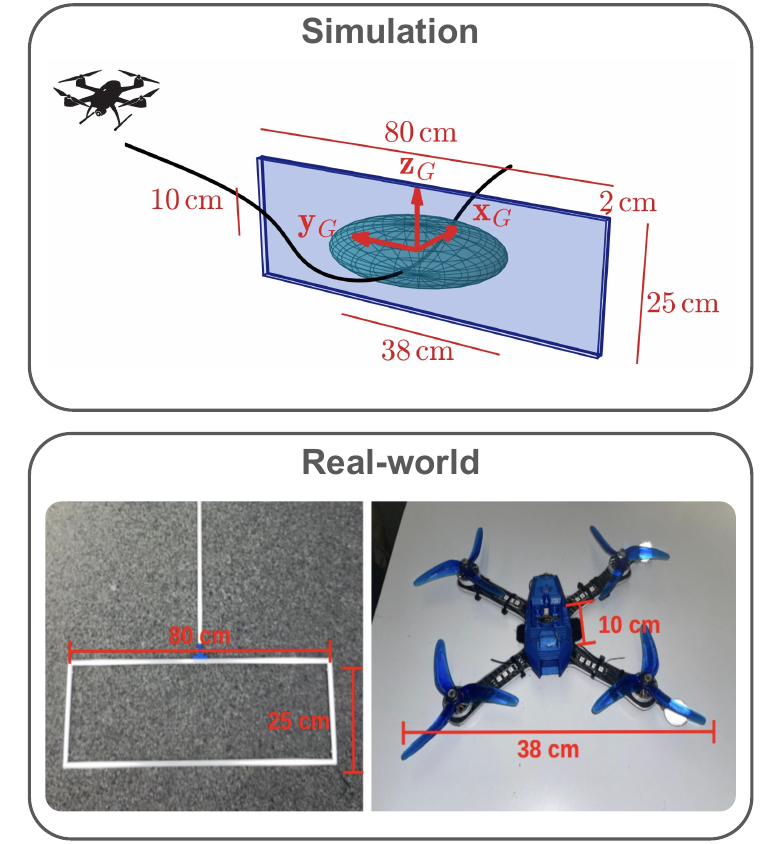}
            \caption{}
            \label{task_geometry_A}
        \end{subfigure}
    \end{minipage}
    \hfill
    \begin{minipage}{0.53\linewidth}
        \centering
        \begin{subfigure}{\linewidth}
            \includegraphics[width=\linewidth]{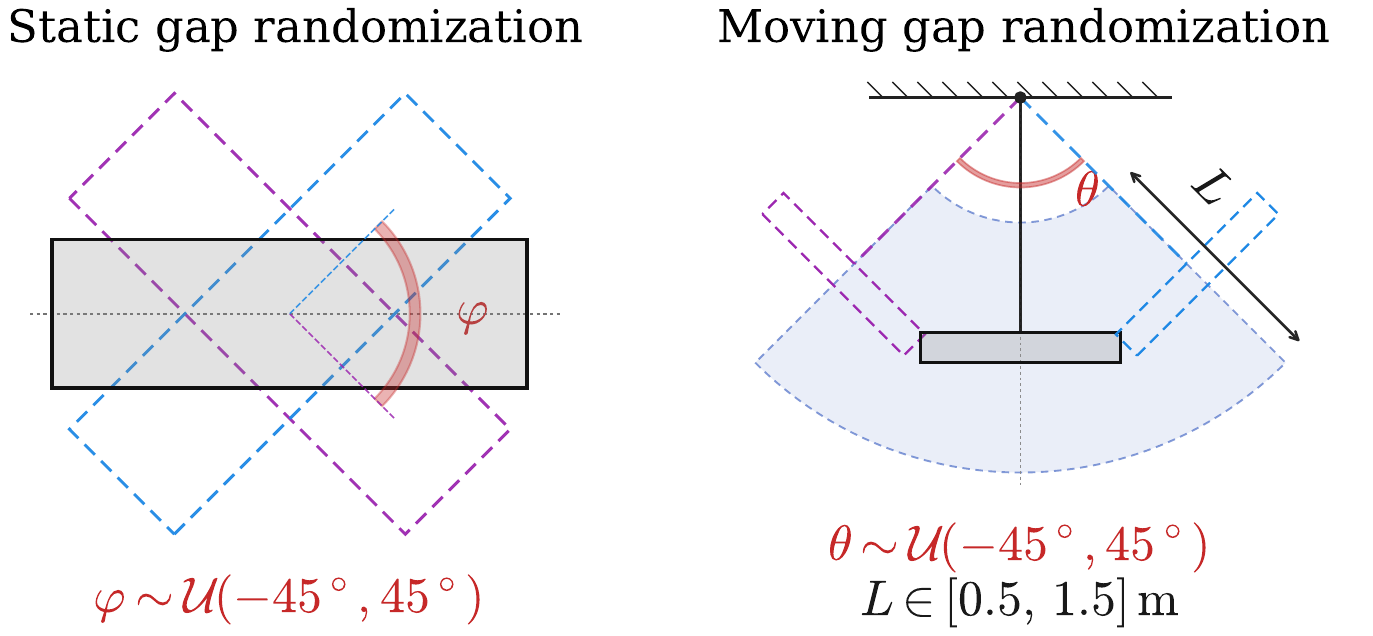}
            \caption{}
            \label{task_geometry_BC}
        \end{subfigure}

        \begin{subfigure}{\linewidth}
            \includegraphics[width=\linewidth]{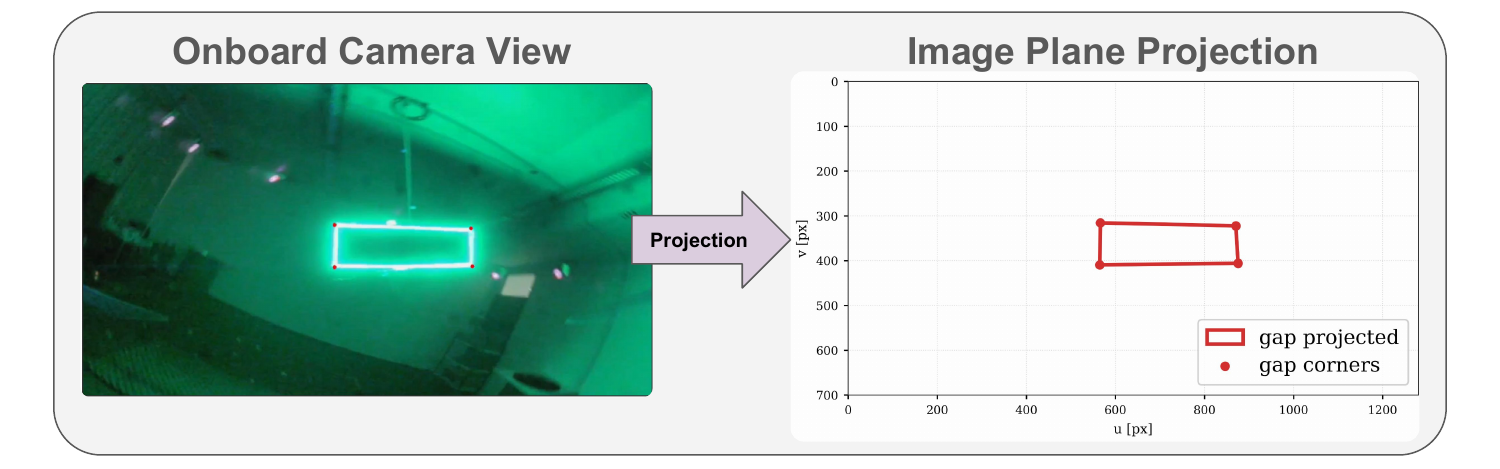}
            \caption{}
            \label{task_geometry_D}
        \end{subfigure}
    \end{minipage}
    \caption{Task geometry. (a) dimensions, (b) static and moving gap randomization, (c) drone camera view (left) and projected gap on image plane (right).
    }
    \label{fig:dims}
    \vspace{-0.8cm}
\end{figure}

\section{Methodology}
\label{sec:methodology}

\subsection{Problem Formulation}
\label{sec:problem_formulation}

We consider closed-loop quadrotor traversal through narrow rectangular gaps in two settings: (i) static gaps and (ii) moving gaps with time-varying pose.
We approximate the quadrotor body as an ellipsoid. A rollout is successful if the ellipsoid projection onto the gap plane is fully contained within the rectangular aperture during traversal.

\paragraph{Control Architecture.}
We use a hierarchical control stack: a low-level onboard controller runs at 1000\,Hz and tracks mass-normalized collective thrust and body-rate commands. A learned policy $\pi_\theta$ runs at 60\,Hz and maps observations to commands,
$\mathbf{u}_t = \pi_\theta(\mathbf{o}_t) = \big[T_t\;\; \boldsymbol{\omega}_{\mathrm{des},t}\big]$,
where $T_t\in\mathbb{R}$ is the collective thrust command and $\boldsymbol{\omega}_{\mathrm{des},t}\in\mathbb{R}^3$ are desired body rates expressed in the body frame.
To model command latency, the policy output is applied to the simulated dynamics with a one-step delay, i.e., the dynamics at step $t$ use $\mathbf{u}_{t-1}$ rather than the newly computed $\mathbf{u}_t$.

The quadrotor (approx.\ $10\times 38\times 38$\,cm) must traverse a rectangular gap of size $25\times 80\times 2$\,cm without collision (Fig.~\ref{task_geometry_A}). A gap frame $G$ is centered at the aperture, with $x_G$ normal to the gap plane and $(y_G,z_G)$ aligned with width and height. We consider static gaps, rotated about $x_G$ by $\phi \in [-45^\circ,45^\circ]$, and moving gaps following planar pendulum motion with randomized amplitude and length (Fig.~\ref{task_geometry_BC}), resulting in a time-varying pose $(\mathbf{p}_G(t), \mathbf{R}_G(t))$. The task geometry and observation modalities are illustrated in Fig.~\ref{fig:dims}.

\paragraph{Observation Settings.}
\label{sec:obs_settings}

We consider state-based and visual-feature-based observations.

The state-based observation provides privileged quadrotor and gap information:
\begin{equation}
    \mathbf{o}_{\mathrm{state}}=
    \big[
    \mathbf{p},\, \mathrm{vec}(\mathbf{R}),\, \mathbf{v},\, \boldsymbol{\omega},\,
    \mathbf{p}_{\mathrm{goal}},\, \mathbf{p}_G,\, \mathrm{vec}(\mathbf{R}_G),\,
    \mathrm{vec}(\mathbf{A}_{t})
    \big],
\end{equation}
where $\mathrm{vec}(\cdot)$ flattens rotation matrices, $\mathbf{p}_{\mathrm{goal}}$ denotes a target behind the gap, and $\mathbf{A}_t$ is a queue of past actions. For moving gaps, $\mathbf{p}_{\mathrm{goal}}$ is defined relative to the nominal pose.

\begin{figure}[t]
    \centering
    \includegraphics[width=0.98\linewidth]{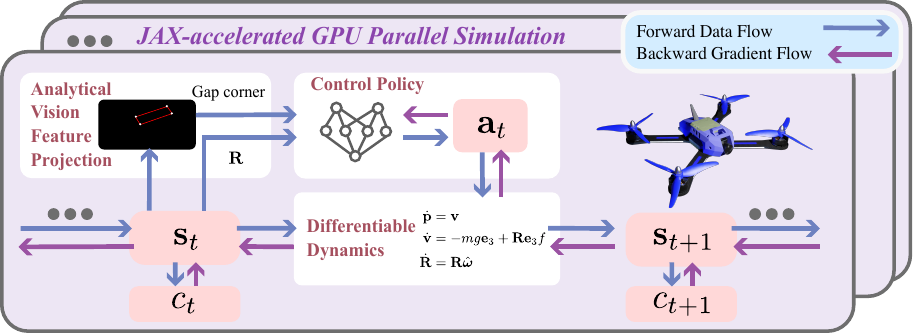}
    \caption{
        Pipeline overview. Policy outputs are propagated through differentiable quadrotor dynamics. Differentiable costs $c_t$ are accumulated over the closed-loop rollout, and gradients are backpropagated through the trajectory to optimize the observation-to-command policy directly.
    }
    \label{fig:diffsim}
\end{figure}

The visual-feature-based observation excludes global pose information and instead uses onboard inertial quantities $(\mathbf{R}, \boldsymbol{\omega})$ together with normalized 2D gap-corner projections (Fig.~\ref{task_geometry_D}) and additional features derived from projected gap corners. These features (e.g., projected center, area, aspect ratio, visibility) are deterministic functions of the corner projections and are augmented with short-term temporal differences. The observation also includes the action history $\mathrm{vec}(\mathbf{A}_t)$. Full details are provided in Appendix~\ref{sec:appendix_observations}.

\subsection{Differentiable Simulation and Objective}
\label{sec:diffsim}

We optimize $\pi_\theta$ via BPTT through differentiable closed-loop rollouts, with $\mathbf{x}_{t+1}=f(\mathbf{x}_t,\mathbf{u}_t)$ implemented via RK4 integration (Appendix~\ref{sec:appendix_dynamics}). The objective is
\begin{equation}
    \mathcal{J}(\theta)=\mathbb{E}_{\tau \sim \pi_\theta}\!\left[\sum_{t=0}^{T} c_t\right],
\end{equation}
with gradients computed through the unrolled computation graph (Fig.~\ref{fig:diffsim}).

We use a differentiable per-step cost
\begin{equation}
    c_t = c_{\mathrm{progress}} +
    c_{\mathrm{centering}} +
    c_{\mathrm{near\_centering}} +
    c_{\mathrm{alignment}} +
    c_{\mathrm{collision}} +
    c_{\mathrm{stabilization}} +
    c_{\mathrm{cam}},
\end{equation}
and define $r_t=-\Delta t\,c_t$ so maximizing return corresponds to minimizing the time-integral of the cost.
Intuitively, the terms enforce: (i) progress through the gap, (ii) centering and refined centering near the gap plane,
(iii) attitude alignment with the gap frame at traversal, (iv) collision avoidance using a smooth differentiable penalty,
(v) regularization of velocities/accelerations/actions, and (vi) (vision-only) maintaining the gap in the camera field of view.
Most of the terms are modulated as a function of the signed distance to the gap plane, increasing geometric precision near traversal, while relaxing stabilization terms during the crossing phase.
For moving gaps, the plane and its signed distance are time-varying; consequently, centering, alignment, and smooth collision terms remain coupled to the instantaneous aperture pose throughout the rollout.
Full formulations and weights are in Appendix~\ref{sec:appendix_reward}.

\subsection{Training Setup}
\label{sec:training_setup}

Episodes start $2$\,m before the gap with random position offsets $\Delta\mathbf{p}\sim \mathcal{U}([-1,1]^3)$\,m and velocity perturbations $\mathbf{v}_0,\boldsymbol{\omega}_0\sim \mathcal{U}([-0.1,0.1]^3)$. Roll and pitch are initialized at zero, and yaw is randomized in $[-40^\circ,40^\circ]$. Episodes terminate upon collision or after $T=2.5$\,s. Domain randomization reduces the sim-to-real gap and includes gap rotation (static), pendulum phase, amplitude, and length (moving), perception--actuation delay, thrust scaling, and (for vision) gap-corner noise and dropouts (see Appendix~\ref{sec:app_domain_randomization}). Policies are trained via BPTT with batched rollouts using a 2-layer MLP (width 512 for state-based and 1024 for visual-feature-based observations), optimized with Adam and a warmup--cosine schedule. Training uses 100 parallel environments for up to 20k epochs, selecting the best validation checkpoint (full hyperparameters in Appendix~\ref{sec:app_training_details}).

\section{Experiments and Results}
\label{sec:experiments_results}

\begin{figure}[t]
    \centering
    \begin{minipage}[t]{0.33\textwidth}
        \vspace{0pt}
        \centering

        \begin{subfigure}[t]{\linewidth}
            \vspace{0pt}
            \includegraphics[width=\linewidth]{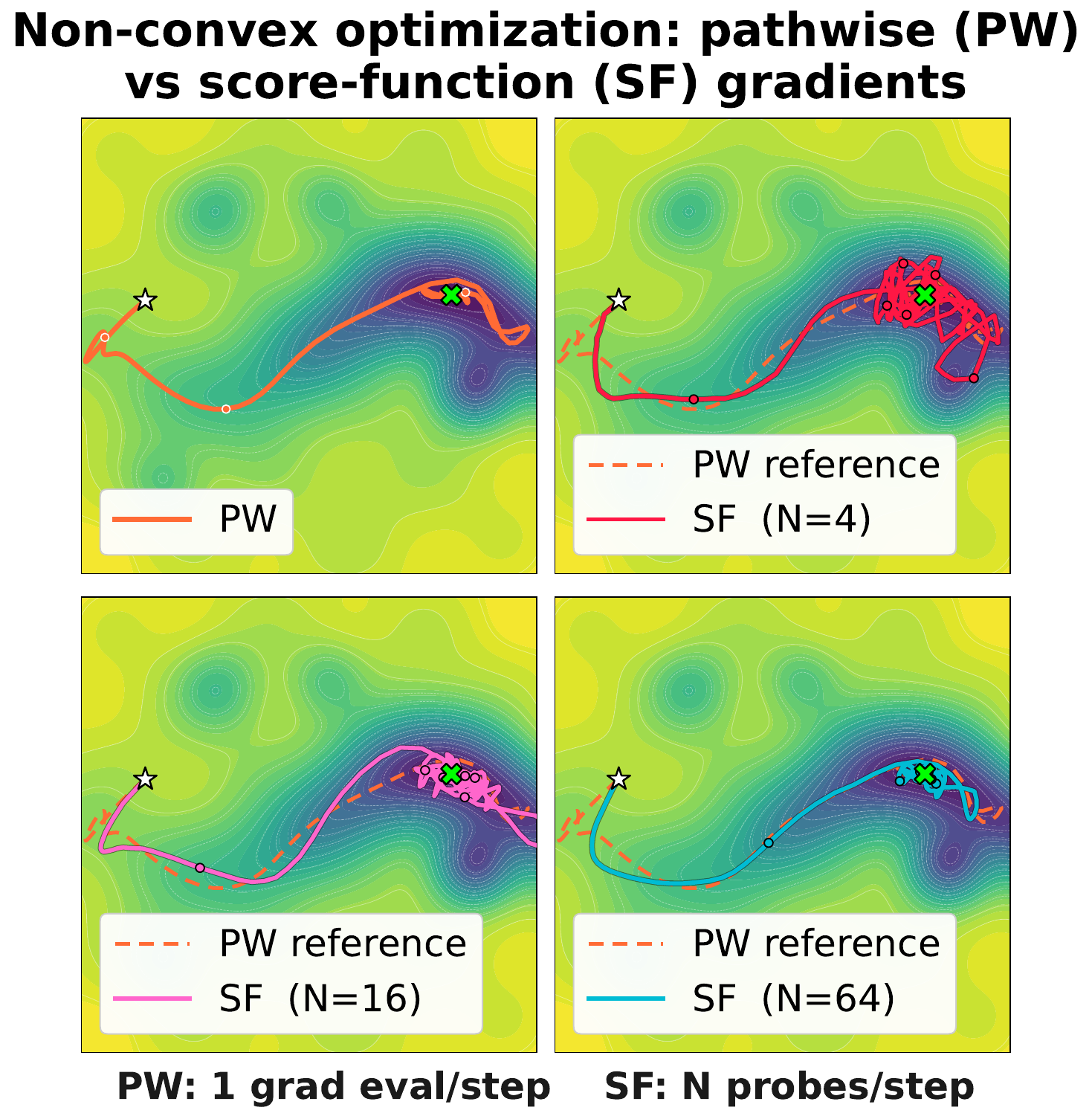}
            \caption{}
            \label{pw_vs_fs_A}
        \end{subfigure}

        \begin{subfigure}[t]{\linewidth}
            \vspace{0pt}
            \includegraphics[width=\linewidth]{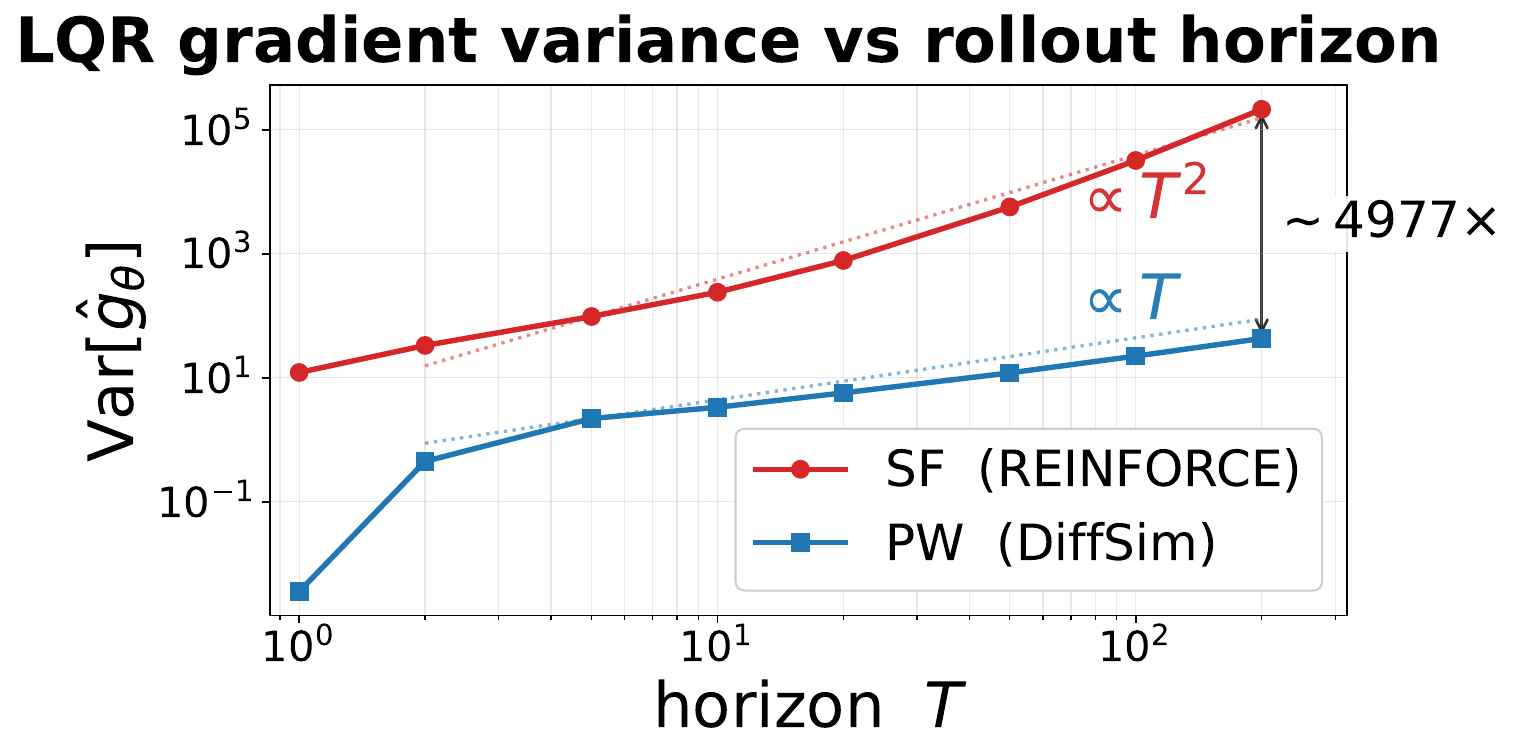}
            \caption{}
            \label{pw_vs_fs_B}
        \end{subfigure}
    \end{minipage}%
    \begin{minipage}[t]{0.63\textwidth}
        \vspace{0pt}
        \centering
        \begin{subfigure}[t]{\linewidth}
            \vspace{0pt}
            \includegraphics[width=\linewidth]{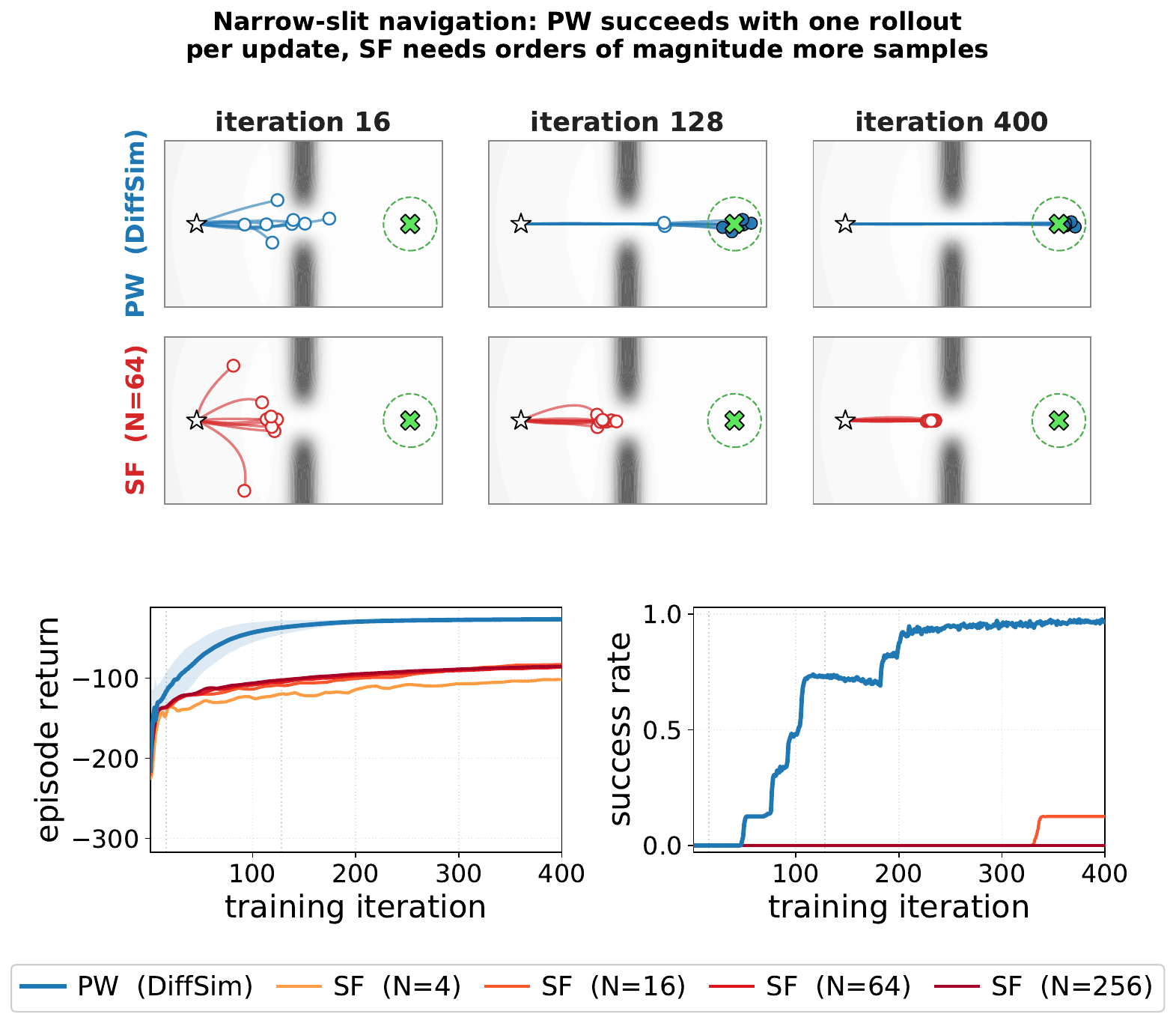}
            \caption{}
            \label{pw_vs_fs_C}
        \end{subfigure}
    \end{minipage}

    \caption{PW vs SF gradient comparison.
        (a) Non-convex 2D landscape and induced gradients.
        (b) Gradient variance as a function of the horizon $T$.
        (c) Slit navigation: episode return and success rate vs training iterations, with example rollouts at different stages.}
    \label{fig:pw-vs-sf}
\end{figure}

\paragraph{Evaluation Setup.}
We evaluate differentiable-simulation training for narrow-gap traversal under full (state-based) and partial (visual-feature-based) observability, considering static and moving gaps. We report success rate, geometric precision at gap crossing, and training efficiency (epochs and wall-clock time), and compare against PPO under varying parallelism. All experiments use a single NVIDIA RTX 4080 SUPER GPU.

\paragraph{Pathwise vs.\ Score-Function Gradients.}
Before turning to the quadrotor, we isolate the key property motivating differentiable simulation: the gap between pathwise (PW) and score-function (SF) gradient estimators under matched settings. The comparison is illustrated in Fig.~\ref{fig:pw-vs-sf}.
PW closely follows the analytic gradient on a non-convex landscape (Fig.~\ref{pw_vs_fs_A}), while SF deteriorates with limited samples; variance scales as $\mathcal{O}(T)$ for PW vs.\ $\mathcal{O}(T^2)$ for SF on an LQR system (Fig.~\ref{pw_vs_fs_B}); and on a neural slit-navigation task (Fig.~\ref{pw_vs_fs_C}), PW solves the problem with a single rollout per update, whereas SF fails with up to 4,096 rollouts per update. This sample ratio refers only to the controlled slit task, not to quadrotor PPO transitions.

\paragraph{Efficiency and Performance.}
\label{par:efficiency_performance}

\begin{wraptable}[10]{r}{0.40\columnwidth}
    \centering
    \small
    \setlength{\tabcolsep}{5pt}

    \begin{tabular}{@{}lccc@{}}
        \toprule
        \textbf{Task} & \textbf{Succ.} & \textbf{Err.} & \textbf{Epochs} \\
        \midrule
        \rowcolor{gray!10}
        State-static  & 100.0          & 1.0           & 500             \\
        Vision-static & 98.9           & 2.4           & 3500            \\
        State-moving  & 100.0          & 2.4           & 1500            \\
        Vision-moving & 99.3           & 2.3           & 4000            \\
        \bottomrule
    \end{tabular}

    \caption{
        Task metrics for differentiable-simulation policies. Success (\%), crossing error (cm), and convergence epochs.
    }
    \label{tab:combined_results}
\end{wraptable}

\begin{figure}[t]
    \centering
    \includegraphics[width=\linewidth]{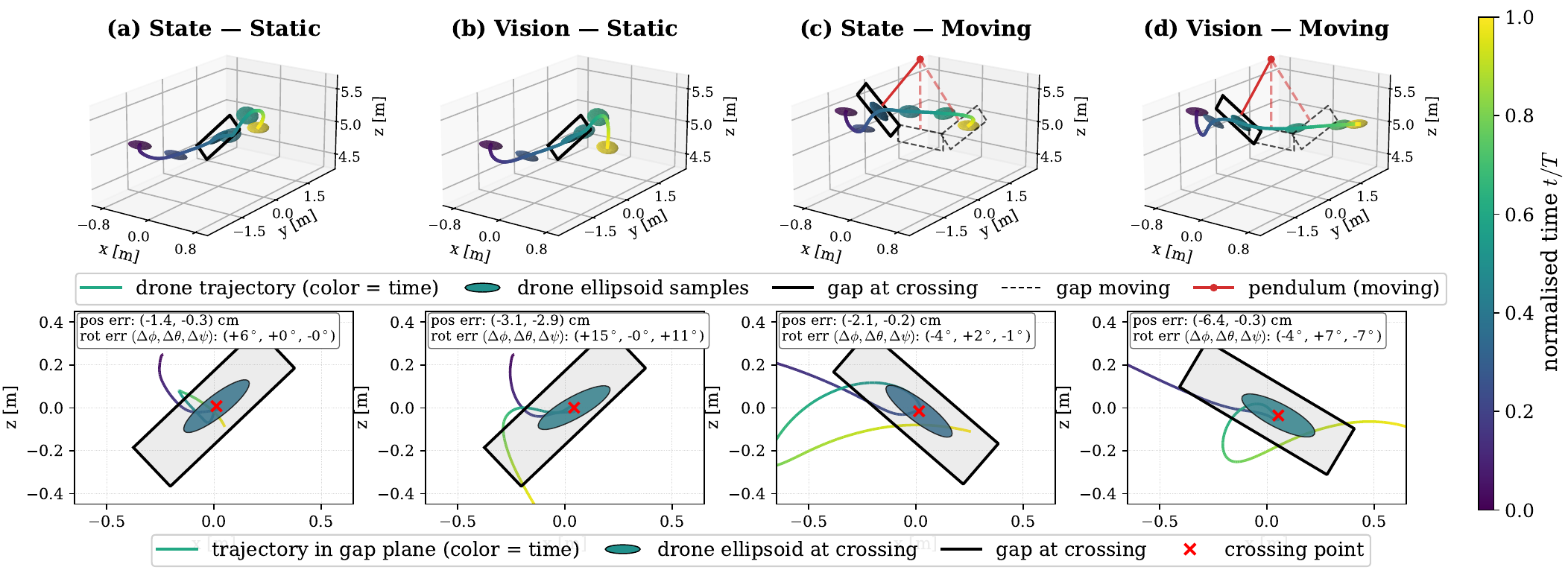}
    \caption{
        Successful rollouts for static and moving gaps under state- and vision-based observations.
    }
    \label{fig:variant_rollouts_placeholder}
\end{figure}

In the quadrotor setting, we evaluate convergence speed together with success and geometric accuracy at gap crossing. Convergence is defined as the first evaluation checkpoint (every 500 epochs) starting a streak of at least three consecutive evaluations with success $\ge 95\%$. At crossing, we measure success rate, Euclidean error to the gap center, in-plane errors along the gap-frame axes ($y_G$/$z_G$), and roll, pitch, and yaw errors. Table~\ref{tab:combined_results} summarizes success, Euclidean crossing error (cm), and convergence epochs for all task variants; per-axis position and attitude errors are omitted for readability. Differentiable simulation achieves near-perfect success and centimeter-level accuracy across all tasks. Performance degrades slightly in the vision-based setting, and convergence requires more optimization steps, reflecting partial observability effects discussed below. Qualitative rollouts illustrating the learned behaviors across task variants are shown in Fig.~\ref{fig:variant_rollouts_placeholder}.

\paragraph{Static-to-Moving Transfer.}
Without fine-tuning, the submitted state-static and vision-static checkpoints achieved 31.5\% and 6.6\% success, respectively, over 1,000 moving-gap rollouts under the Table~\ref{tab:combined_results} protocol, compared with 100\% and 99.3\% after moving-task training. Static-only training therefore did not transfer reliably in our setting. The state and vision tasks also differ in gap-pose distribution, so this experiment does not isolate timing. Unlike the fully observed state policy, the vision policy loses gap feedback near crossing; transfer may therefore depend on gap motion relative to clearance during this blind interval, consistent with its larger drop.

\paragraph{Transfer and Real-World Deployment.}
\begin{wraptable}[7]{r}{0.40\columnwidth}
    \centering
    \small
    \setlength{\tabcolsep}{5pt}

    \begin{tabular}{@{}lcc@{}}
        \toprule
        \textbf{Setting} & \textbf{Succ. (\%)} & \textbf{Trials} \\
        \midrule
        Train sim.       & 99.3                & 1\,000          \\
        Hi-fi sim.       & 81.0                & 100             \\
        Real-world       & 80.0                & 10              \\
        \bottomrule
    \end{tabular}

    \caption{
        Sim-to-real transfer results.
    }
    \label{tab:sim2real}
\end{wraptable}
The visual-feature-based moving-gap policy achieves 99.3\% success over
1\,000 simulated rollouts under the training distribution
(Table~\ref{tab:combined_results}). To assess how this checkpoint degrades
beyond the training simulator, we additionally evaluate it in a higher-fidelity
simulator~\citep{aljalbout2025reality} and on a real quadrotor with onboard sensing
(Table~\ref{tab:sim2real}). These transfer evaluations are not used for
training or checkpoint selection. Success decreases with increasing realism,
primarily due to dynamics mismatch, perception noise, and discrepancies in the
moving-gap mechanism, but the same policy still performs closed-loop traversal
on real hardware. During real-world deployment, LED-illuminated gate edges are detected in
onboard images and converted into gap-corner features
(Fig.~\ref{fig:sim2real}, bottom; see Appendix~\ref{sec:appendix_perception} for details).
The resulting visual features are fed directly to the learned
policy, which outputs collective-thrust and body-rate commands
during real-world moving-gap traversal.
We recorded twelve flights; two had onboard logs consistent with success but lacked external video and were excluded. Of the ten verifiable flights, 8/10 succeeded, with a 95\% Wilson interval of 49--94\%. Failures 03/08 started at $(fwd,lat,vert)=(-1.73,+1.11,-1.10)$ and $(-2.04,-0.30,-0.41)$\,m relative to the nominal gap. Trial 03 was comparable to successful initializations, whereas trial 08 started substantially higher; ten scored trials are insufficient to separate initialization effects from other causes. Both failures fell within the successful-trial ranges of gap-motion amplitude and angular rate at policy start ($35.8$--$42.1^\circ$; $-68.5$ to $+78.3^\circ$/s) and of gap angle and angular rate at crossing ($-27.3$ to $+22.4^\circ$; $-94.3$ to $+96.3^\circ$/s). Both crossed low and hit the bottom edge, with estimated penetrations of 4.6 and 9.3\,cm. Together with the drop from 99.3\% in training simulation to 81\% in high-fidelity simulation, this is consistent with residual dynamics or tracking mismatch affecting vertical accuracy, although the available trials do not isolate it from perception or initialization effects.

\begin{figure}[t]
    \centering
    \includegraphics[width=\linewidth]{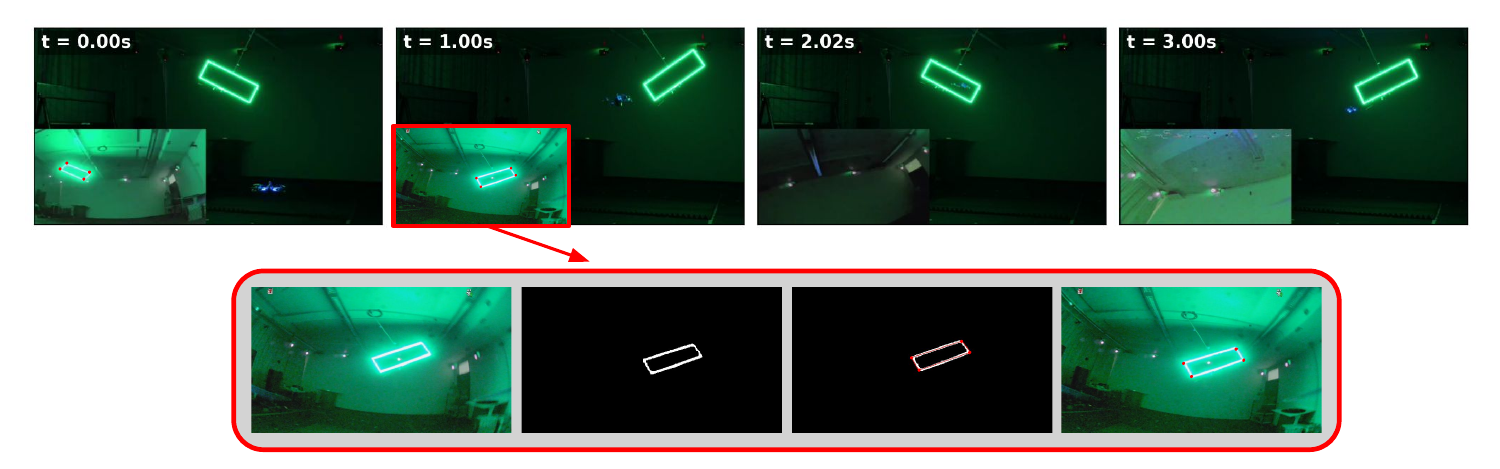}
    \caption{
        Real-world deployment of the vision-based moving-gap policy.
        Top: third-person flight sequence; each frame includes an inset with the onboard camera view.
        Bottom: perception pipeline on one frame — raw image, HSV segmentation mask, quadrilateral fitting, and detected corners.
    }
    \label{fig:sim2real}
\end{figure}

\begin{table}[t]
    \centering
        \footnotesize
        \setlength{\tabcolsep}{3pt}
        \begin{tabular}{llccccccc}
        \toprule
            \textbf{Task} & \textbf{Method}
            & \multicolumn{3}{c}{\textbf{Resources}}
            & \multicolumn{3}{c}{\textbf{Speedup}}
            & \textbf{Conv.} \\
            \cmidrule(lr){3-5} \cmidrule(lr){6-8}
            & & \textbf{Transitions} & \textbf{Updates} & \textbf{Time (s)}
            & \textbf{Trans.} & \textbf{Updates} & \textbf{Time} & \\
        \midrule
        \multirow{2}{*}{Static}
                & BPTT & 75\,M & 500 & $38.3{\pm}0.7$ & \multirow{2}{*}{$2.8\times$} & \multirow{2}{*}{$2.24{\times}10^3$} & \multirow{2}{*}{$6.4\times$} & 5/5 \\
                & PPO & $210{\pm}34$\,M & $1.12{\pm}0.18$\,M & $246{\pm}39$ & & & & 5/5 \\
        \midrule
        \multirow{2}{*}{Moving}
                & BPTT & $135{\pm}34$\,M & $900{\pm}224$ & $67.5{\pm}16.6$ & \multirow{2}{*}{$19.0\times$} & \multirow{2}{*}{$15.2{\times}10^3$} & \multirow{2}{*}{$45.5\times$} & 5/5 \\
                & PPO & $2.57{\pm}1.10$\,B & $13.7{\pm}5.7$\,M & $3073{\pm}1295$ & & & & 4/5 \\
        \bottomrule
    \end{tabular}
    \vspace{0.3cm}
    \caption{
                Matched 1,000-environment state-task budgets on an RTX 4080 SUPER over five seeds. Transition, optimizer-update, and wall-clock speedups are PPO/BPTT ratios computed from the reported means. PPO moving-task statistics average the four converged seeds; seed 2 was censored at 30k epochs (4.5\,B transitions, 24\,M updates, 5,377\,s).
    }
    \label{tab:ppo_convergence}
\end{table}

\paragraph{Comparison with PPO.}

\begin{figure}[t]
    \centering
    \includegraphics[width=\linewidth]{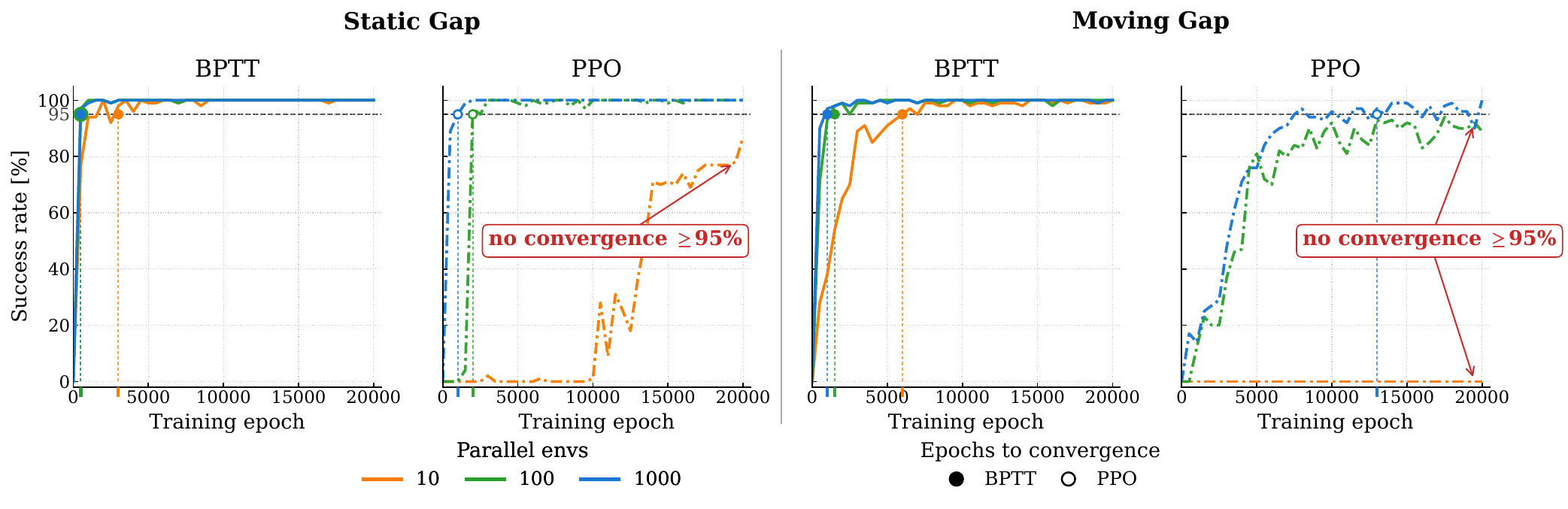}
    \caption{BPTT vs tuned PPO across 10, 100 and 1000 parallel environments (evaluation every 500 epochs over 100 rollouts).}
    \label{fig:ppo_compare}
    \vspace{-0.4cm}
\end{figure}

We use PPO as a representative on-policy SF baseline to compare full-horizon BPTT with standard model-free policy optimization, not as a pathwise baseline. The PPO baseline uses the same actor and 150-step horizon, but tuned objective weights and a curriculum; we therefore do not attribute the entire pipeline difference to the estimator. At 1,000 environments both methods process 150k transitions per epoch, but BPTT takes one full-rollout gradient step whereas PPO takes 800 minibatch steps, so epochs are not compute-equivalent. Table~\ref{tab:ppo_convergence} instead reports transitions, optimizer updates, hardware, wall-clock time, and five seeds. On moving gaps, converged-seed means show that PPO uses $19\times$ more transitions and $45.5\times$ more training time than BPTT. Figure~\ref{fig:ppo_compare} retains the submitted parallelism sweep.

\begin{figure}[t]
    \centering

    \includegraphics[width=0.8\linewidth]{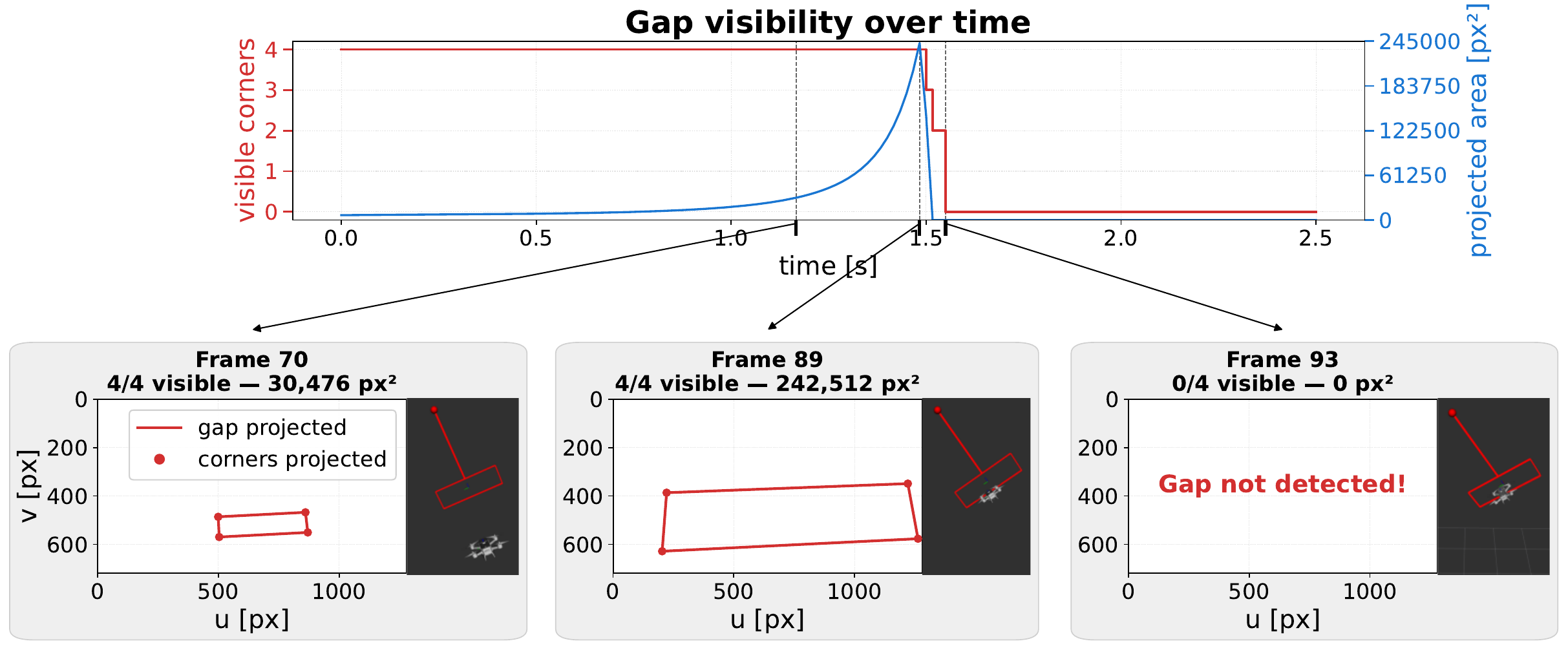}
    \caption{
        Gap visibility over one episode. Top: visible gap corners (red) and projected gap area (blue) over time; dashed lines mark the snapshots below. Bottom: onboard image plane (left) and third-person view (right) at three instants, showing the gap exiting the field of view.}
    \label{fig:partial_obs}
\end{figure}

\begin{wraptable}{r}{0.40\columnwidth}
    \centering
    \footnotesize
    \setlength{\tabcolsep}{3pt}
    \begin{tabular*}{\linewidth}{@{\extracolsep{\fill}}ccc@{}}
        \toprule
        $K$ & \textbf{Success (\%)} & \textbf{Convergence epoch} \\
        \midrule
        1 & 99.5 & 6,500 \\
        2 & 98.8 & 4,500 \\
        4 & 99.3 & 4,000 \\
        8 & 98.9 & 3,500 \\
        \bottomrule
    \end{tabular*}
    \caption{Vision-moving performance as the temporal window length $K$ varies.}
    \label{tab:partial_obs_k}
\end{wraptable}
\paragraph{Partial Observability in Visual-Feature-Based Policies.}
\label{par:partial_obs}
Invalid corners are sentinel-marked and their derived geometry is masked, while earlier valid frames remain in the window. Once no valid corners remain, the policy has no gap-relative visual information and uses $R$, $\omega$, and the four-command queue. The MLP window provides a short explicit history rather than predicting future gap pose (Fig.~\ref{fig:partial_obs}). Table~\ref{tab:partial_obs_k} reports the vision-moving results as $K$ varies. Larger $K$ converged earlier while final success varied by 0.7 points; $K=1$ removes visual-frame history but retains $R$, $\omega$, and the command queue. We did not compare an RNN, which may help during longer dropout.

\section{Discussion and Limitations}
\label{sec:discussion_limitations}
Differentiable simulation achieves near-perfect simulated success and centimeter-level accuracy across the tested static and planar pendulum-driven moving gaps. The controlled studies show informative pathwise gradients in this smooth, locally stable regime, while the complete BPTT pipeline uses fewer transitions and less training time than tuned PPO. Because PPO uses tuned weights and a curriculum, the entire pipeline difference cannot be attributed to the estimator alone.

\paragraph{Limitations.}
Training costs use privileged simulator geometry and collision data, and the visual input consists of engineered geometric features rather than raw images. The physical setup relies on LED-based corners and evaluates only planar pendulum motion. Natural openings, abrupt or non-periodic motion, severe or prolonged visual dropout, hard contact, and long-horizon instability remain untested. The ten scored flights yield broad statistical uncertainty, and their two failures do not isolate dynamics/tracking mismatch from perception or initialization effects. Finally, the method requires differentiable components and incurs BPTT memory and compute costs that scale with horizon and fidelity.

\paragraph{Future work.}
Reducing model mismatch through learnable residual dynamics or more accurate differentiable models is a natural direction. Marker-free perception, recurrent policies for longer dropout, and evaluation under broader gap motions would extend the present controlled proof of concept. Short-horizon pathwise methods with bootstrap values may also be preferable to full-horizon BPTT for longer tasks.

Overall, differentiable simulation offers a strong training paradigm for high-precision flight, enabling fast and stable optimization in regimes where model-free RL is limited by gradient variance
and sample complexity.

\clearpage
\acknowledgments{This work was supported by the European Union’s Horizon Europe Research and Innovation Programme under grant agreement No. 101120732 (AUTOASSESS) and the European Research Council (ERC) under grant agreement No. 864042 (AGILEFLIGHT).}

\bibliography{example}
\clearpage
\appendix
\section{Supplementary Materials}

\subsection{Gradient Analysis and Properties of Pathwise Estimators}
\label{app:toys}

This section complements Fig.~\ref{fig:pw-vs-sf} by providing a deeper
analysis of the structural differences between pathwise (PW) and
score-function (SF) gradient estimators. We detail the experimental
setup used to probe their behaviour and present two additional analyses
that do not fit in the main text: (i) the dependence of estimator
variance on the closed-loop stability margin in a controlled dynamical
system (App.~\ref{app:lqr}), and (ii) the behaviour of the
score-function estimator as a function of the number of parallel
rollouts in a narrow-passage navigation task (App.~\ref{app:slit}).
All experiments are implemented in PyTorch and reproduced by the scripts
released with the codebase.
We use full-horizon APG/BPTT rather than proposing a new gradient estimator or optimizer. Full-horizon differentiation fits our smooth 150-step rollout; short-horizon pathwise methods with a bootstrap value may suit longer horizons. We did not compare other pathwise methods.

\subsection{Common Setup and Hyperparameters}
\label{app:setup}

\paragraph{Estimators.}
We compare two estimators of
$\nabla_\theta \mathbb{E}[\mathcal{L}(\theta)]$ under matched optimizer,
learning rate, and number of gradient updates:
\begin{itemize}
  \item \textbf{Pathwise (PW)}:
  $\hat g_{\text{PW}} =
    \nabla_\theta \mathcal{L}(\tau(\theta, \varepsilon))$,
  obtained via reverse-mode autodiff through the deterministic rollout.

  \item \textbf{Score-function (SF)}:
  $\hat g_{\text{SF}} =
    \tfrac{1}{N}\sum_{i=1}^{N} \big(R(\tau_i) - b\big)\,
    \nabla_\theta \log p_\theta(\tau_i)$,
  where $b$ is a leave-one-out baseline and $N$ is the number of
  independent rollouts per update.
\end{itemize}
The SF estimator is implemented as REINFORCE to isolate estimator-level
effects; while methods such as PPO may improve optimization stability,
they do not alter the underlying credit assignment mechanism.

\paragraph{Controlled dynamical system.}
We consider a 1D linear system
$s_{t+1} = a s_t + b u_t + w_t$ with
$w_t \sim \mathcal{N}(0,\sigma_w^2)$, a linear stochastic policy
$u_t = \theta s_t + \sigma_\pi \varepsilon_t$, and quadratic loss
$\mathcal{L} = \sum_{t=0}^{T-1}(s_t^2 + \lambda u_t^2)$.
Default parameters are:
$a=1.0,\ b=1.0,\ \lambda=0.1,\ \sigma_w=0.1,\ \sigma_\pi=0.2,\ s_0=1$,
evaluated at $\theta=-0.5$ unless stated otherwise.

\paragraph{Narrow-passage navigation.}
We study a planar control problem with dynamics
$x_{t+1} = x_t + \Delta t\,u_t$, $\|u_t\| \le u_{\max}$,
$\Delta t=0.1$, $u_{\max}=1$, and horizon $T=40$.
The start and goal are $(-1,0)$ and $(1,0)$ respectively.
Both estimators optimize the same reward:
\begin{equation}
  r(x) = -\|x - x_g\|^2
  \;-\; \alpha \exp\!\Big(-\tfrac{x^2}{2\sigma_w^2}\Big)
  \Big(1 - \exp\!\Big(-\tfrac{y^2}{2\sigma_{\text{slit}}^2}\Big)\Big),
\end{equation}
with $\alpha=20$, $\sigma_w=0.05$, and
$\sigma_{\text{slit}}=0.15$.
The policy is an MLP $2 \to 16 \to 2$ with $\tanh$ activations.
Both methods use Adam with learning rate $5\times10^{-3}$ and 8 seeds.
PW uses a single deterministic rollout per update, while SF is evaluated
over $N \in \{4,16,64,256,1024,4096\}$ rollouts.

\paragraph{Success criterion.}
At each iteration we evaluate 100 deterministic rollouts with
Gaussian perturbations $\mathcal{N}(0,0.15^2 I)$ applied to the start.
A rollout is successful if it crosses $x=0.3$ and its final position lies
within a radius $0.25$ of the goal.

\subsection{Variance Scaling: Horizon and Stability Regime}
\label{app:lqr}

The $\mathcal{O}(T)$ versus $\mathcal{O}(T^2)$ scaling reported in the
main text can be derived from a variance decomposition of the
estimators. For the SF estimator,
\begin{equation}
  \hat g_{\text{SF}} =
  \Big(\sum_{t=0}^{T-1}(s_t^2 + \lambda u_t^2)\Big)
  \sum_{t=0}^{T-1} \nabla_\theta \log \pi_\theta(u_t \mid s_t),
\end{equation}
a trajectory-level return that grows linearly with $T$ multiplies a sum
of $T$ score terms whose residual variance scales as $\Theta(T)$,
leading to an overall $\Theta(T^2)$ scaling. In contrast, the PW
estimator propagates sensitivities through the dynamics; in the stable
regime the resulting Jacobian norm remains bounded by a geometric
series, yielding $\Theta(T)$ scaling.

\paragraph{Stability ablation.}
\begin{figure}[t]
  \centering
  \includegraphics[width=0.95\linewidth]{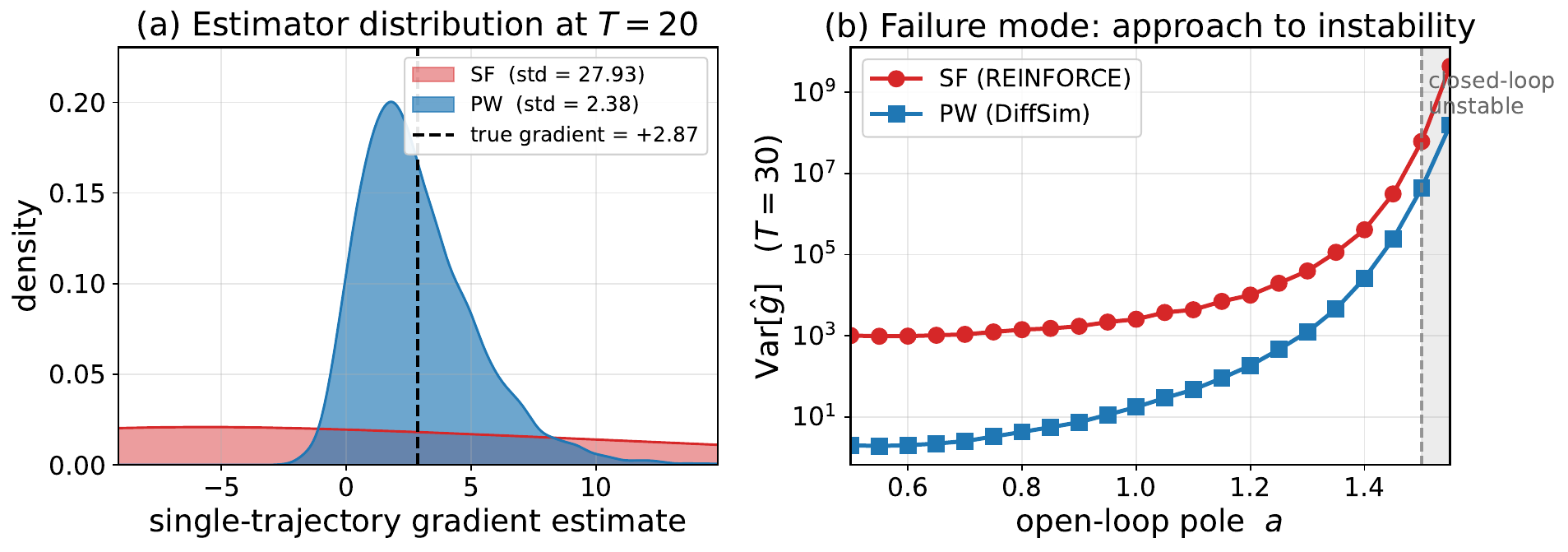}
  \caption{\textbf{LQR estimator analysis (extended).}
  \textit{(a)} Distribution of single-trajectory gradient estimates at
  horizon $T=20$, evaluated at $\theta=-0.5$. Both estimators are
  unbiased around the true gradient (black dashed line), but the
  pathwise (PW) estimator concentrates more than an order of magnitude
  more tightly than the score-function (SF) one ($\mathrm{std}\,2.38$
  vs.\ $27.93$ in our setup, i.e.\ a per-trajectory variance ratio of
  $\sim\!140\times$).
  \textit{(b)} Stability ablation: sweeping the open-loop pole
  $a \in [0.5, 1.55]$ at fixed $\theta=-0.5$ shifts the closed-loop
  pole $a + b\theta$ across unity at $a=1.5$ (shaded region). Both
  estimators diverge as the system approaches instability, but PW
  retains a $>\!10^{2}\!\times$ variance advantage up to the
  bifurcation, after which both saturate. This is the regime in which
  PW loses its asymptotic edge — and the regime that the quadrotor
  gap-traversal task avoids by design.}
  \label{fig:lqr-extended}
\end{figure}
Fig.~\ref{fig:lqr-extended} sweeps $a \in [0.5,1.55]$ at fixed
$\theta=-0.5$, such that the closed-loop pole $a + b\theta$ crosses unity
at $a=1.5$. Both estimators degrade near instability, but PW retains a
large advantage up to the bifurcation. This highlights a key property:
the pathwise estimator is not universally superior; its advantage
emerges in regimes with smooth, locally stable, and moderately
long-horizon dynamics. This analysis diagnoses the smooth, locally stable regime used in our 150-step quadrotor rollouts; it does not establish a new stability boundary.

\subsection{Signal vs. Variance in Narrow-Passage Problems}
\label{app:slit}

\paragraph{Parallelism does not compensate for missing signal.}
Fig.~\ref{pw_vs_fs_C} reports $N \in \{4,16,64,256\}$; under the same protocol, additional runs at $N=1{,}024$ and $4{,}096$ also fail.
All SF configurations plateau at similar success rates
($\approx 0.10$–$0.12$), indicating that increasing $N$ reduces estimator
variance but does not increase the underlying signal, which remains
near-zero before the passage is discovered. Under Gaussian exploration,
the probability of sampling a successful trajectory is effectively
negligible, making the reward signal uninformative regardless of $N$.

\paragraph{Trajectory-level interpretation.}
The trajectory snapshots make this behaviour explicit. The PW estimator
updates the policy using the analytic spatial gradient
$\partial r / \partial x$ even before successful trajectories are
observed, progressively shaping the trajectory toward the feasible
manifold. In contrast, the SF estimator produces trajectories that move
in the correct direction but remain outside the success region, reflecting
a weak credit signal dominated by sampling noise.

\subsection{Observation Details}
\label{sec:appendix_observations}

\subsubsection{State-Based Observations}

In the state-based setting, the observation vector consists of the full
quadrotor state, the gap pose, the goal position, and a short history of
past actions:
\begin{equation}
\mathbf{o}_\text{state} =
\begin{bmatrix}
\mathbf{p} \\
\mathrm{vec}(\mathbf{R}) \\
\mathbf{v} \\
\boldsymbol{\omega} \\
\mathbf{p}_\text{goal} \\
\mathbf{p}_G \\
\mathrm{vec}(\mathbf{R}_G) \\
\mathrm{vec}(\mathbf{A}_t)
\end{bmatrix}
\;\in\;\mathbb{R}^{\,3+9+3+3+3+3+9+N_a\cdot 4},
\end{equation}
where $\mathbf{p}$, $\mathbf{R}\!\in\!SO(3)$, $\mathbf{v}$ and
$\boldsymbol{\omega}$ are the quadrotor position, orientation,
linear velocity and body-frame angular velocity, $(\mathbf{p}_G,\mathbf{R}_G)$
is the gap pose, and $\mathbf{p}_\text{goal}$ is the post-gap waypoint.
The operator $\mathrm{vec}(\cdot)$ flattens a rotation matrix into a
$9$-dimensional row-major vector.

\paragraph{Action history.}
$\mathbf{A}_t \!\in\! \mathbb{R}^{N_a \times 4}$ is a FIFO queue of the last
$N_a$ commanded actions, each of the form
$\mathbf{a} = [c_\text{thrust},\,\omega_x^\text{ref},\,\omega_y^\text{ref},\,\omega_z^\text{ref}]$
(collective thrust and body-rate references). The queue is initialized with
the hovering action and is updated, at every control step, by removing the
oldest entry and appending the latest action. Its length is set to cover the
maximum perception–actuation delay,
\begin{equation}
N_a \;=\; \left\lfloor \frac{\texttt{delay\_max}}{\texttt{dt}} \right\rfloor + 2,
\end{equation}
where in our implementation the per-episode delay is uniformly sampled in
$[\texttt{delay}-0.005,\,\texttt{delay}+0.005]\,\text{s}$, so that
$\texttt{delay\_max}=\texttt{delay}+0.005$. With nominal values $\texttt{dt} = \tfrac{1}{60}\,\text{s}$ and $\texttt{delay}=0.035\,\text{s}$
(hence $\texttt{delay\_max}=0.040\,\text{s}$) we obtain $N_a = 4$, and
$\mathrm{vec}(\mathbf{A}_t)$ contributes $4\times 4 = 16$ additional scalars
to the observation. This buffer allows the policy to compensate the
randomized control latency and acts as a smoothness regularizer through the
action-rate penalty.

\subsubsection{Visual-Feature-Based Observations}

The visual-feature-based observation combines onboard inertial measurements
with features derived from the projection of the gap corners over a short
temporal window of $K$ frames (with an optional stride of \texttt{skip\_frames}
between consecutive frames).

\paragraph{Projected Corner Features.}
The four 3D corners of the rectangular gap are defined as
\begin{equation}
\mathbf{c}_i =
\mathbf{p}_G
\pm \tfrac{w}{2}\,\mathbf{u}_G
\pm \tfrac{h}{2}\,\mathbf{v}_G,
\qquad i=1,\dots,4,
\end{equation}
where $\mathbf{u}_G$ and $\mathbf{v}_G$ are the width and height unit vectors
of the gap frame. Each corner is projected into the image plane through a
(double-sphere) camera model, given the body pose $(\mathbf{p},\mathbf{R})$:
\begin{equation}
\tilde{\mathbf{z}}_i = \Pi\!\left(\mathbf{R}_{CB}\,{}_B\mathbf{t}_{C c_i}\right),
\qquad
\mathbf{z}_i = \frac{2\,\tilde{\mathbf{z}}_i}{(W,H)} - \mathbf{1},
\end{equation}
so that valid pixels are normalized to $[-1,1]^2$. During training, before
normalization, an independent uniform pixel noise
$\boldsymbol{\epsilon}_i \sim \mathcal{U}(-\epsilon_{\max},\epsilon_{\max})^2$
(with $\epsilon_{\max}=10$ px) is added to each corner to simulate detection
uncertainty (see domain-randomization section). Corners falling outside the
field of view are flagged invalid and replaced by a sentinel value
$\mathbf{z}_i \leftarrow -0.999\cdot\mathbf{1}$. The four
corners are then sorted in a consistent clockwise order starting from the
top-right corner, removing permutation ambiguity.

\paragraph{Derived Geometric Features.}
From the (sorted, possibly invalid) projected corners we deterministically
compute, \emph{per frame}:
\begin{itemize}
  \item the projected center
        $\mathbf{c} = \tfrac{1}{4}\sum_{i=1}^4 \mathbf{z}_i$,
        set to $-0.999$ if any corner is invalid;
  \item the normalized projected area
        \begin{equation}
        A = \frac{1}{4}\cdot\tfrac{1}{2}
        \left|\sum_{i=1}^{4} \big(x_i y_{i\oplus 1} - x_{i\oplus 1} y_i\big)\right|,
        \end{equation}
        with $\oplus$ denoting addition modulo $4$ (Shoelace formula),
        normalized so that the full image area equals $1$;
  \item the projected axes
        $\mathbf{u}_\text{img} = \widehat{\mathbf{r}-\boldsymbol{\ell}}$,
        $\mathbf{v}_\text{img} = \widehat{\mathbf{t}-\mathbf{b}}$,
        where $\mathbf{r},\boldsymbol{\ell},\mathbf{t},\mathbf{b}$ are the
        midpoints of the right, left, top and bottom edges of the projected
        quadrilateral;
  \item the projected aspect ratio
        $\rho = \lVert\mathbf{r}-\boldsymbol{\ell}\rVert /
                \lVert\mathbf{t}-\mathbf{b}\rVert$;
  \item the visibility score
        $\nu = \tfrac{1}{4}\sum_{i=1}^4 \mathbf{1}[i\text{ valid}] \in [0,1]$;
  \item the horizontal-alignment scalar
        $\alpha = \mathbf{u}_\text{img}\!\cdot\![1,0]^\top$.
\end{itemize}
All geometric features collapse to $0$ when at least one corner is invalid.
Earlier valid frames remain in the temporal window after an invalid detection. Once every retained frame is invalid, no gap-relative visual information remains and the policy acts from $\mathbf{R}$, $\boldsymbol{\omega}$, and the four-command queue.

\paragraph{Temporal Window and Deltas.}
The features above are computed over a sliding window of the last $K$ frames, producing the stacked tensors
$\mathbf{Z}\!\in\!\mathbb{R}^{K\times 4\times 2}$,
$\mathbf{C}\!\in\!\mathbb{R}^{K\times 2}$,
$\mathbf{A}\!\in\!\mathbb{R}^{K}$, etc.
Motion cues are encoded as differences between the last two frames of the
window:
\begin{equation}
\Delta\mathbf{c} = \mathbf{C}_K - \mathbf{C}_{K-1},
\qquad
\Delta A     = \mathbf{A}_K - \mathbf{A}_{K-1}.
\end{equation}

\paragraph{Final Observation Vector.}
Concatenating the inertial part, the stacked visual features, the temporal
deltas and the action history $\mathrm{vec}(\mathbf{A}_t)$, the final
observation is
\begin{equation}
\mathbf{o}_\text{vision} =
\begin{bmatrix}
\mathrm{vec}(\mathbf{R}) \\[2pt]
\boldsymbol{\omega} \\[2pt]
\mathrm{vec}(\mathbf{Z}) \\[2pt]
\mathrm{vec}(\mathbf{C}) \\[2pt]
\mathbf{A} \\[2pt]
\boldsymbol{\rho} \\[2pt]
\boldsymbol{\nu} \\[2pt]
\mathrm{vec}(\mathbf{U}_\text{img}) \\[2pt]
\mathrm{vec}(\mathbf{V}_\text{img}) \\[2pt]
\boldsymbol{\alpha} \\[2pt]
\Delta\mathbf{c} \\[2pt]
\Delta A \\[2pt]
\mathrm{vec}(\mathbf{A}_t)
\end{bmatrix}
\;\in\;\mathbb{R}^{\,9 + 3 + K(8+2+1+1+1+2+2+1) + 2 + 1 + n_a\,d_a},
\end{equation}
with $K=4$ in our experiments.

\subsection{Perception Pipeline}
\label{sec:appendix_perception}

To extract gap corner observations in real-world experiments, we employ a lightweight detection pipeline based on LED illumination. The gate is equipped with a high-intensity green LED strip along its boundary. Due to the high brightness, the LEDs appear as saturated white regions in the camera image. Images are converted to HSV and thresholded with $H \in [0, 100]$, $S \in [0, 100]$, $V \in [200, 255]$ to isolate these bright regions into a binary mask. A morphological closing operation (11$\times$11 kernel) is applied to fill small gaps in the mask, followed by connected component filtering to discard regions with area below 500\,px.

\begin{figure}[t]
\centering

\includegraphics[width=0.24\linewidth]{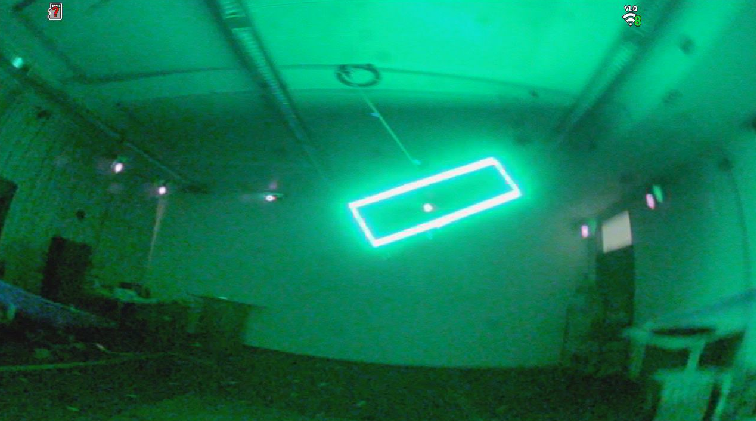}
\hfill
\includegraphics[width=0.24\linewidth]{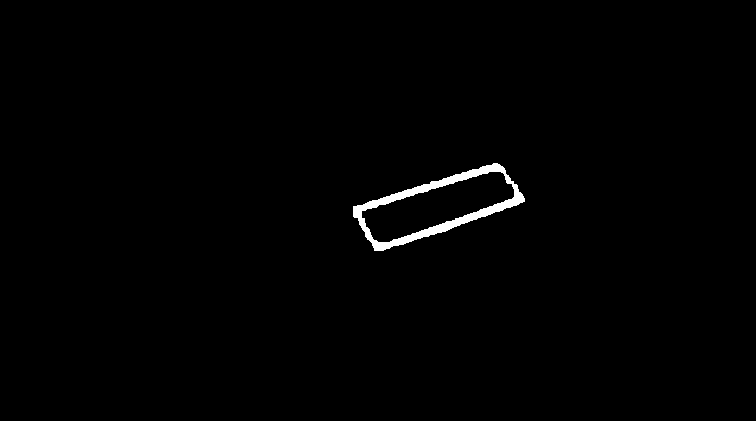}
\hfill
\includegraphics[width=0.24\linewidth]{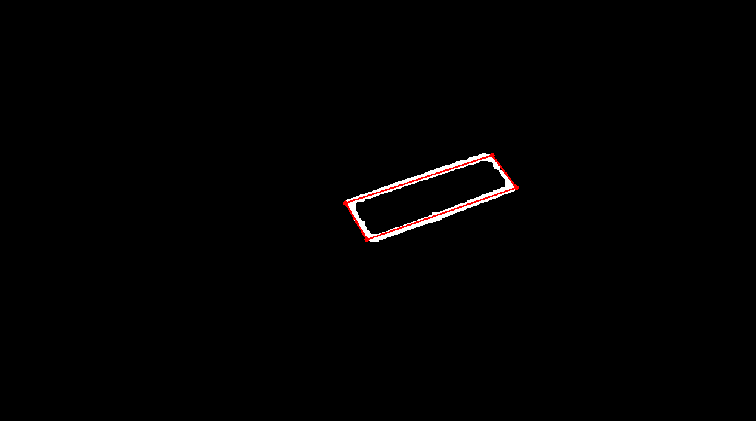}
\hfill
\includegraphics[width=0.24\linewidth]{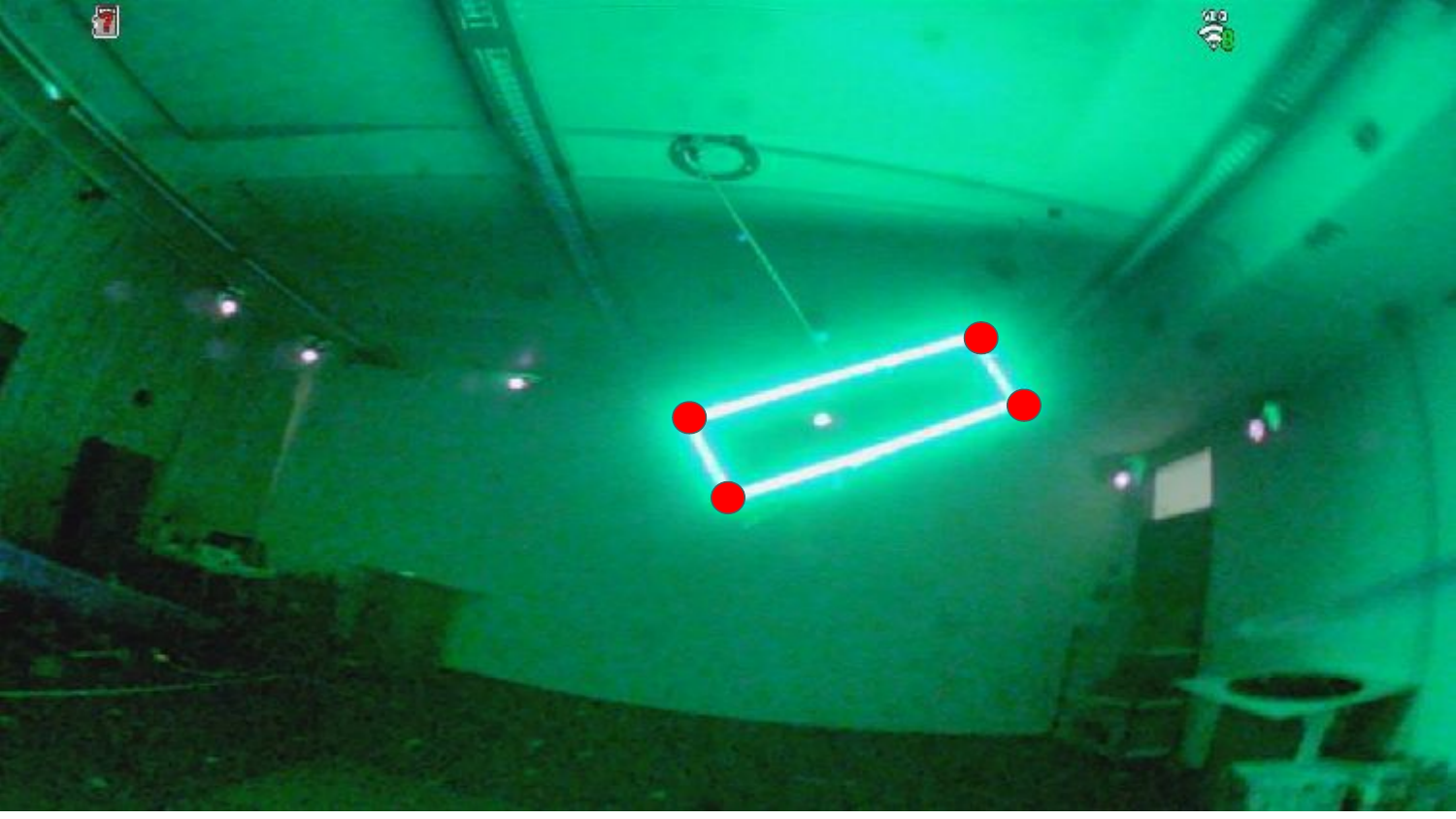}

\caption{
Gate detection pipeline used for extracting gap corners in real-world experiments.
From left to right: raw image, segmentation mask after HSV thresholding and noise removal, quadrilateral fitting, and final detected corners (red dots).
}
\label{fig:perception_pipeline}
\end{figure}

The remaining region is assumed to correspond to the gap and is approximated with a quadrilateral via contour fitting. The resulting four vertices are interpreted as the gap corners. When multiple candidates are detected, the one closest to the image center is selected. For temporal consistency, detected corners are matched across frames and sorted in a fixed clockwise order. These corners directly define the visual observation features used by the policy. The perception pipeline is illustrated in Fig.~\ref{fig:perception_pipeline}.

This simplified perception pipeline isolates the control problem and enables controlled evaluation, although it relies on strong assumptions on the visual structure of the environment.

\subsection{Reward Details}
\label{sec:appendix_reward}

We provide the full formulation of the trajectory cost introduced in Sec.~\ref{sec:diffsim}.
All terms are differentiable with respect to states and actions and are combined to optimize both geometric precision and dynamic feasibility.

\begin{figure}[t]
\centering
\includegraphics[width=\linewidth]{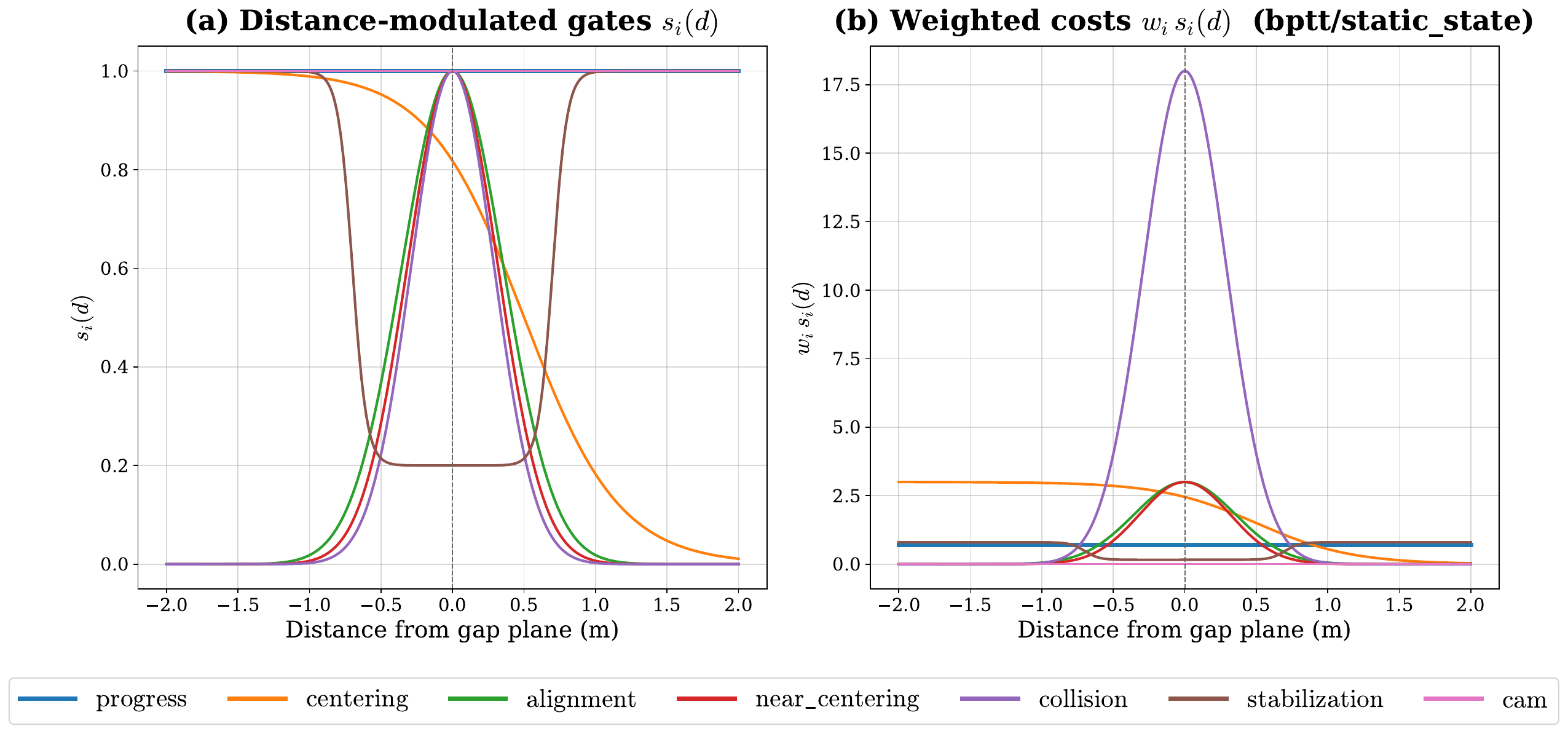}
\caption{
Distance modulation overview.
(a) Shaping functions $s_i(d)$ used for distance-dependent weighting.
(b) Resulting weighted, distance-modulated costs for the state-static task.
}
\label{fig:distance_modulation_overview}
\end{figure}

\paragraph{Distance-dependent weighting.}
Let $\mathbf{n}_G$ denote the gap normal and $\mathbf{p}_G$ the gap center. The signed distance is
\begin{equation}
d(\mathbf{p}) = (\mathbf{p} - \mathbf{p}_G)^\top \mathbf{n}_G .
\end{equation}
For each cost component $c_i$, we use a distance-modulated weight
\begin{equation}
w_i(d) \;=\; w_i \, s_i(d),
\label{eq:weight_modulation}
\end{equation}
where $w_i \ge 0$ is a task-specific base weight and $s_i(d)$ is a smooth shaping function.
Table~\ref{tab:reward_weights} reports the base weights $w_i$ used for each task configuration, while Fig.~\ref{fig:distance_modulation_overview} visualizes the shaping functions $s_i(d)$ and weighted costs $w_i\,s_i(d)$ for BPTT static-state task.
For terms that are not distance-modulated we set $s_i(d)\equiv 1$ (progress and camera alignment).

\paragraph{Smooth penalty.}

All geometric costs use a SmoothL1 penalty to ensure stable gradients:
\begin{equation}
\text{SmoothL1}(x) =
\begin{cases}
\frac{1}{2}x^2, & |x| < 1, \\
|x| - \frac{1}{2}, & \text{otherwise}.
\end{cases}
\end{equation}
Scaled as $\text{SmoothL1}(\alpha x)/\alpha$ to control sharpness.

\paragraph{Progress cost.}

Encourages the quadrotor to reach a goal $\mathbf{p}_{\text{goal}}$ behind the gap, placed at a distance of $2$\,m along the gap normal. For moving gaps, $\mathbf{p}_{\text{goal}}$ is defined relative to the nominal gap pose ($\theta = 0$). The progress cost equation is:
\begin{equation}
c_{\text{progress}} =
w_{\text{progress}}\, s_{\text{progress}}(d)\;
\frac{\text{SmoothL1}(\alpha \|\mathbf{p} - \mathbf{p}_{\text{goal}}\|)}{\alpha},
\qquad s_{\text{progress}}(d)\equiv 1.
\end{equation}

\paragraph{Centering costs.}

Let
\begin{equation}
e_u = (\mathbf{p} - \mathbf{p}_G)^\top \mathbf{u}_G, \quad
e_v = (\mathbf{p} - \mathbf{p}_G)^\top \mathbf{v}_G.
\end{equation}

The centering cost penalizes deviation from the gap center:
\begin{equation}
c_{\text{centering}} =
w_{\text{centering}}\, s_{\text{centering}}(d)\;
\Big(
\text{SmoothL1}(e_u) + \beta \,\text{SmoothL1}(e_v)
\Big).
\end{equation}

A second near-gap term enforces higher precision close to traversal:
\begin{equation}
c_{\text{near\_centering}} =
w_{\text{near\_centering}}\, s_{\text{near\_centering}}(d)\;
\Big(
\text{SmoothL1}(e_u) + \beta \,\text{SmoothL1}(e_v)
\Big).
\end{equation}

Their distance shaping functions differ across the two terms:
\begin{equation}
s_{\text{centering}}(d) = \sigma\!\big(k(\delta - d)\big),
\qquad
s_{\text{near\_centering}}(d) = \exp\!\big(-\lambda_{\text{near}} d^2\big),
\label{eq:centering_shapes}
\end{equation}
where $\sigma(\cdot)$ is the logistic sigmoid. In our implementation, $k=3$ and $\delta=0.5$, matching
$\sigma(-3(d-0.5))$, and $\lambda_{\text{near}}=5$.

We use $\beta=3$ to weight the vertical centering term more strongly, reflecting the tighter clearance margin along the gap height (narrow $z_G$ aperture).

\paragraph{Alignment cost.}

The relative rotation is
\begin{equation}
\mathbf{R}_{\text{rel}} = \mathbf{R}_G^\top \mathbf{R}.
\end{equation}

We extract the roll, pitch, and yaw angles $(\theta_{\text{roll}},\theta_{\text{pitch}},\theta_{\text{yaw}})$ from $\mathbf{R}_{\text{rel}}$ using a standard Euler-angle decomposition and normalize them to $[-\pi,\pi]$.

The cost penalizes roll, pitch, and yaw misalignment:
\begin{equation}
c_{\text{alignment}} =
w_{\text{alignment}}\, s_{\text{alignment}}(d)\;
\left(
\theta_{\text{yaw}}^2 + 2\theta_{\text{pitch}}^2 + 2\theta_{\text{roll}}^2
\right).
\end{equation}

We weight pitch and roll more strongly than yaw, since pitch/roll directly affect the projected clearance through the aperture, while yaw is less critical for collision-free traversal.

\paragraph{Collision cost.}
\begin{wrapfigure}{r}{0.48\columnwidth}
\centering
\includegraphics[width=\linewidth]{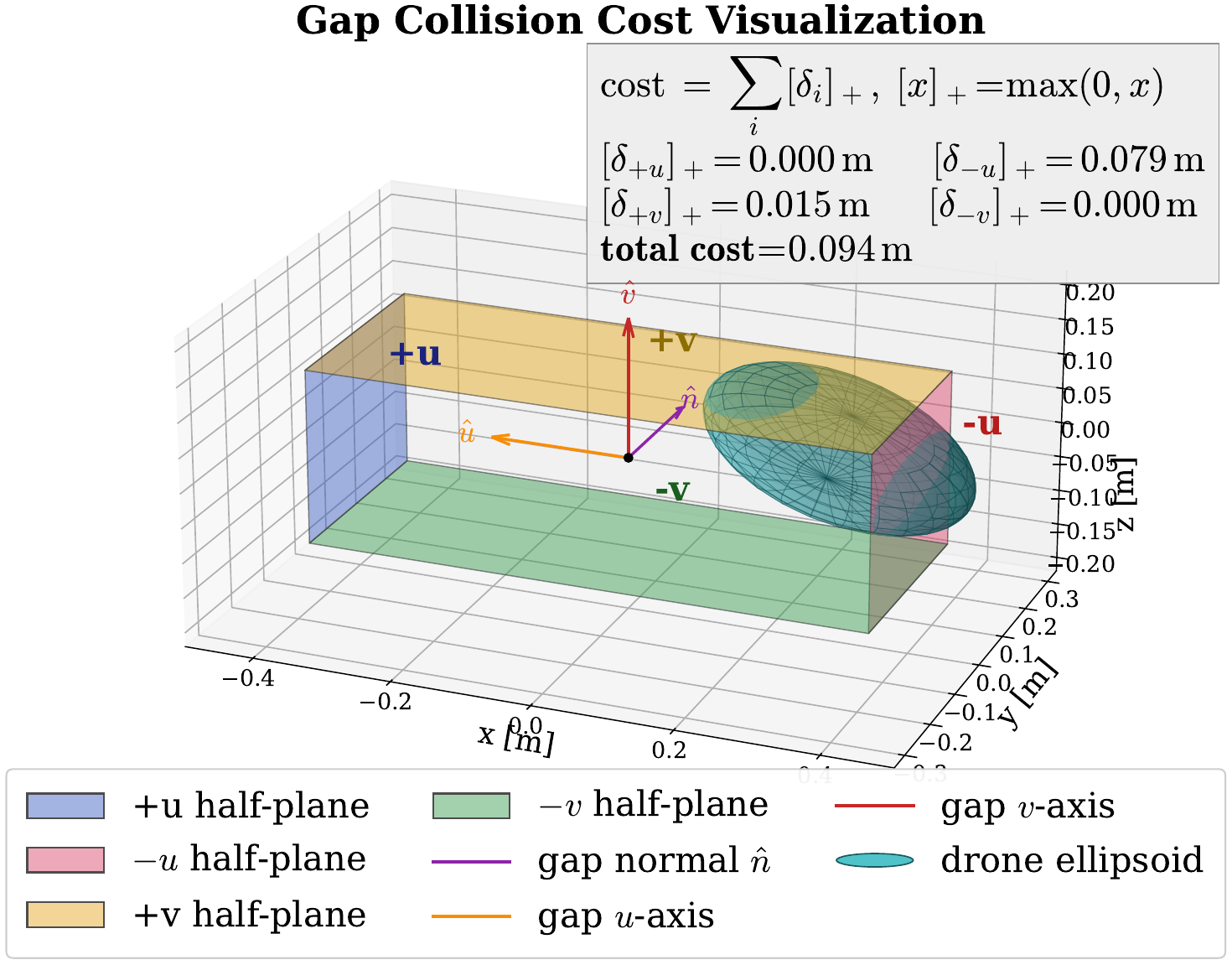}
\caption{Collision cost as a function of penetration depth.}
\label{fig:collision_barrier}
\end{wrapfigure}

To enforce strict geometric feasibility during traversal, the quadrotor is approximated as a rigid ellipsoid and the gap aperture is modeled as a set of planar half-space constraints.

Let $\mathbf{p} \in \mathbb{R}^3$ be the quadrotor position and $\mathbf{R} \in SO(3)$ its orientation.
The ellipsoidal body is defined by a shape matrix $\mathbf{Q} \in \mathbb{R}^{3 \times 3}$ describing its principal axes.

The gap is represented in its local frame $G$ by four planar constraints corresponding to the gap boundaries.
Let $\mathbf{u}_G$ and $\mathbf{v}_G$ denote the width and height directions of the gap.
We define the outward normals as
\begin{equation}
\mathbf{A} =
\begin{bmatrix}
\mathbf{u}_G & \mathbf{v}_G & -\mathbf{u}_G & -\mathbf{v}_G
\end{bmatrix}^\top
\end{equation}

and the corresponding offsets $\mathbf{b} \in \mathbb{R}^4$ such that the feasible region satisfies
\begin{equation}
\mathbf{A} \mathbf{p} \leq \mathbf{b}.
\end{equation}

To account for the finite size of the quadrotor, we compute a conservative bound on the ellipsoid extent along each constraint normal.
The projected radius of the ellipsoid along a direction $\mathbf{n}$ is approximated by $\|\mathbf{n}^\top \mathbf{R} \mathbf{Q}\|$.

The resulting penetration vector is defined as
\begin{equation}
\mathbf{c}_{\mathrm{coll}} =
\max\!\Big(
\|\mathbf{A}\mathbf{R}\mathbf{Q}\| + \mathbf{A}\mathbf{p} - \mathbf{b}, \, 0
\Big),
\end{equation}
where the maximum is applied element-wise.

This quantity represents the violation of each half-space constraint after accounting for the quadrotor geometry.

The collision cost is then defined as
\begin{equation}
c_{\text{collision}} =
w_{\text{collision}}\, s_{\text{collision}}(d)\;
\sum_{i=1}^{4} \mathbf{c}_{\mathrm{coll},i},
\qquad s_{\text{collision}}(d)=\exp(-\lambda_{\text{coll}} d^2).
\end{equation}
where $s_{\mathrm{collision}}(d)$ increases the penalty near the gap plane.

This formulation provides a smooth and differentiable approximation of collision constraints, enabling gradient-based optimization while enforcing that the quadrotor remains within the gap aperture at traversal.
The geometric interpretation of collision and alignment cost is shown in Fig.~\ref{fig:collision_barrier}.

\paragraph{Stabilization cost.}

To regularize the dynamics, we penalize velocity, angular velocity, acceleration, control effort:
\begin{equation}
c_{\text{stabilization}} =
w_{\text{stabilization}}\, s_{\text{stabilization}}(d)\;
\left(
w_{\text{vel}}\|\mathbf{v}\| + w_{\boldsymbol{\omega}}\|\boldsymbol{\omega}\| + w_{\text{acc}}\|\mathbf{a}\| + w_{\text{action}}\|\mathbf{u} - \mathbf{u}_{\text{hover}}\|
\right).
\end{equation}

We downweight stabilization near the gap plane using a smooth sigmoid window with a non-zero floor to allow aggressive maneuvers during crossing:
\begin{equation}
s_{\text{stabilization}}(d)
=
1 - (1-s_{\min})\Big(\sigma\!\big(k(d+\delta)\big)-\sigma\!\big(k(d-\delta)\big)\Big),
\label{eq:stab_shape}
\end{equation}
where $\sigma(\cdot)$ is the logistic sigmoid, $\delta$ controls the half-width of the window, and $k$ its steepness.
In our implementation, $s_{\min}=0.2$, $k=20$, and $\delta=0.7$, so that $s_{\text{stab}}(d)\approx 1$ far from the gap plane and reaches a minimum of $s_{\min}$ inside the window. We fix the stabilization weight to $w_{\text{stabilization}}=1$ and the stabilization sub-weights to $w_{\text{vel}}=0.1$, $w_{\boldsymbol{\omega}}=0.1$, $w_{\text{acc}}=0.1$, and $w_{\text{action}}=0.5$ across all tasks.

\paragraph{Camera alignment cost (vision only).}

Encourages the quadrotor to keep the gap in view:
\begin{equation}
c_{\text{cam}} =
w_{\text{cam}}\, s_{\text{cam}}(d)\;
\left(
\frac{\arccos(\mathbf{e}_x^\top \mathbf{d})}{\pi}
\right)^2,
\qquad s_{\text{cam}}(d)\equiv 1.
\end{equation}
where $\mathbf{d}$ is the normalized direction toward the gap.

\begin{table}[t]
\centering
\setlength{\tabcolsep}{4pt}
\begin{tabular}{lcccccc}
\toprule
\textbf{Task} & $w_{\text{progress}}$ & $w_{\text{centering}}$ & $w_{\text{near\_cent.}}$ & $w_{\text{alignment}}$ & $w_{\text{collision}}$ & $w_{\text{cam}}$ \\
\midrule
State-Static   & 0.7 & 3.0 & 3.0 & 3.0  & 18.0 & 0.0 \\
Vision-Static  & 0.7 & 6.0 & 5.0 & 3.0  & 22.0 & 0.7 \\
State-Moving   & 0.5 & 3.0 & 5.0 & 6.0  & 27.0 & 0.0 \\
Vision-Moving  & 0.5 & 5.0 & 5.0 & 6.0  & 30.0 & 0.7 \\
\bottomrule
\end{tabular}
\caption{
Task-specific base weights $w_i$ used for each cost component.
Distance-modulated weights are defined in Eq.~\eqref{eq:weight_modulation}.}
\label{tab:reward_weights}
\end{table}

\subsection{Quadrotor Dynamics}
\label{sec:appendix_dynamics}

The quadrotor state is defined as
\begin{equation}
\mathbf{x} = \big[\mathbf{p},\; \mathbf{v},\; \mathbf{R},\; \boldsymbol{\omega}\big],
\end{equation}
where $\mathbf{p}, \mathbf{v} \in \mathbb{R}^3$ are position and velocity
in the world frame, $\mathbf{R} \in SO(3)$ is the rotation from body to world frame,
and $\boldsymbol{\omega} \in \mathbb{R}^3$ is the body angular velocity. The policy outputs a collective thrust command $T$ and desired body rates
$\boldsymbol{\omega}_{\mathrm{des}}$. The low-level controller tracks these
commands at 1000\,Hz; in simulation, body rates are assumed to be tracked
instantaneously, i.e.\ $\boldsymbol{\omega} = \boldsymbol{\omega}_{\mathrm{des}}$.  To emulate actuator
uncertainty, at every step the commanded thrust is perturbed by a
multiplicative uniform noise,
\begin{equation}
T \;\leftarrow\; T\,(1+\epsilon), \qquad \epsilon\sim\mathcal{U}(-0.1,\,0.1),
\end{equation}
before being applied to the dynamics. The translational dynamics are
\begin{equation}
\dot{\mathbf{p}} = \mathbf{v}, \qquad
\dot{\mathbf{v}} = \mathbf{g} + \mathbf{R}
\begin{bmatrix} 0 \\ 0 \\ T/m \end{bmatrix},
\end{equation}
with $\mathbf{g} = [0\;\; 0\;\; {-9.81}]^\top$\,m/s$^2$ and $m = 0.600$\,kg.
Position and velocity are integrated using fourth-order Runge--Kutta (RK4). Orientation is updated via the exponential map (Rodrigues formula).
For $\boldsymbol{\phi} = \boldsymbol{\omega}\,\Delta t$ and
$\theta = \|\boldsymbol{\phi}\|$:
\begin{equation}
\mathbf{R}_{t+1} = \mathbf{R}_t \exp(\widehat{\boldsymbol{\phi}}),
\qquad
\exp(\widehat{\boldsymbol{\phi}}) =
\mathbf{I}
+ \frac{\sin\theta}{\theta}\,\widehat{\boldsymbol{\phi}}
+ \frac{1 - \cos\theta}{\theta^2}\,\widehat{\boldsymbol{\phi}}^{\,2},
\end{equation}
where $\widehat{\cdot}$ denotes the skew-symmetric operator.
This orientation update is exact for constant angular velocity
over the integration step and avoids quaternion normalization issues. The entire discrete-time model is fully differentiable and constitutes
the transition function $\mathbf{x}_{t+1} = f(\mathbf{x}_t, \mathbf{u}_t)$
used for BPTT optimization.

\subsection{Domain Randomization Distributions}
\label{sec:app_domain_randomization}

\paragraph{Initial state randomization.}
Episodes start $2$\,m before the gap with position offsets
$\Delta\mathbf{p} \sim \mathcal{U}([-1,1]^3)$\,m,
velocity perturbations $\mathbf{v}_0, \boldsymbol{\omega}_0 \sim \mathcal{U}([-0.1, 0.1]^3)$,
and yaw randomized in $[-40^\circ, 40^\circ]$.
Roll and pitch are initialized at zero.

\paragraph{Gap randomization.}
For static gaps, the gap is rotated about its normal by an angle
$\phi \sim \mathcal{U}([-45^\circ, 45^\circ])$.
For moving gaps, we sample an amplitude $\theta_{\max}\in[-45^\circ,-20^\circ]\cup[20^\circ,45^\circ]$
and an initial angle $\theta_0\sim \mathcal{U}([-|\theta_{\max}|,|\theta_{\max}|])$.
The initial angular velocity is set using energy conservation,
\begin{equation}
\dot{\theta}_0 = \pm \sqrt{\frac{2g}{L}\left(\cos\theta_0-\cos\theta_{\max}\right)},
\label{eq:theta0dot_paper}
\end{equation}
with random sign, and the pendulum length is randomized as $L \sim \mathcal{U}([0.5,1.5])\;\text{m}$.
We exclude the central interval $[-20^\circ,20^\circ]$ to avoid near-static configurations.

\paragraph{Perception noise.}
To emulate imperfect perception, we perturb the projected image-plane coordinates with additive noise sampled independently for each coordinate of each corner:
$u \leftarrow u + \epsilon$, with $\epsilon \sim \mathcal{U}([-10,10])$ pixels.
Corners that fall outside the field of view are treated as invalid and assigned a fixed sentinel value.

\paragraph{Perception--actuation delay.}
We introduce a stochastic delay to emulate processing and communication latency.
At the beginning of each episode, the delay is sampled as $\tau = 0.035 + \epsilon$, with $\qquad \epsilon \sim \mathcal{U}([-0.005,\,0.005]) \ \text{s}$, and kept constant during the rollout.
The range is chosen based on measured deployment latency.

\paragraph{Action randomization.}
To account for actuation uncertainty and modeling inaccuracies, we introduce multiplicative noise on the collective thrust command.
At each timestep, the thrust output of the policy is perturbed as $T \leftarrow T \cdot (1 + \epsilon)$, with $\quad \epsilon \sim \mathcal{U}([-0.1, 0.1])$, corresponding to up to $\pm 10\%$ variation in the applied thrust.

\subsection{Training Details}
\label{sec:app_training_details}

\paragraph{Policy architecture.}
The policy is a 2-hidden-layer MLP with ReLU activations.
State-based policies use width 512, visual-feature-based policies use width 1024.
Both BPTT and PPO share the same architecture for a given observation setting.
Policy weights are initialized with a uniform distribution scaled by a factor of $0.01$.

\paragraph{BPTT optimization.}
We use Adam with a warmup--cosine learning-rate schedule.
For state-based tasks, the learning rate is warmed up from $10^{-3}$
to $10^{-2}$ and decayed over 800 steps to $5 \times 10^{-4}$.
For visual-feature-based tasks, the learning rate is warmed up from $10^{-3}$
to $5 \times 10^{-4}$ and decayed over 2000 steps to $5 \times 10^{-5}$.
Training runs for up to 20k epochs with evaluation every 500 epochs
over 100 rollouts; the checkpoint with the highest success rate is selected.

\paragraph{PPO hyperparameters.}
PPO uses the same reward components as BPTT but with independently
tuned weights and a reward curriculum (see below).
All tasks share the following hyperparameters:
discount factor $\gamma = 0.99$,
GAE $\lambda = 0.95$,
clip ratio $\varepsilon = 0.2$,
4 update epochs per iteration,
minibatch size of 750 transitions,
value-function coefficient $0.5$,
entropy coefficient $0.02$,
and maximum gradient norm $1.0$.
The learning rate follows a warmup--cosine schedule from
$10^{-4}$ to a peak of $3 \times 10^{-4}$, decaying to $10^{-5}$.
The number of minibatches per epoch is derived as
$\lfloor (N_{\text{envs}} \cdot T) / 750 \rfloor$, where
$T = 150$ steps ($2.5\,\text{s}$ at $60\,\text{Hz}$), ensuring a
consistent minibatch size across the three parallelism levels
we evaluate (10/100/1000).

\paragraph{Reward curriculum.}
Without curriculum learning, PPO fails to learn gap traversal entirely under the
same full objective used by BPTT: the collision, near-centering, and centering
costs penalize any imprecise crossing attempt, causing the policy to avoid the
gap plane altogether and resulting in zero success rate.
To address this, we linearly anneal these weights from their initial to
final values over a task-dependent epoch range (Table~\ref{tab:ppo_curriculum}).
This allows the policy to first learn to approach and cross the gap,
then progressively improve alignment and geometric precision as the
penalties increase, ultimately achieving successful traversal at the
cost of slower convergence.
BPTT uses fixed task weights throughout training and does not require any
curriculum learning, as pathwise gradients provide sufficient signal to optimize
all cost components jointly from the start.
\begin{table}[h]
\centering
\small
\setlength{\tabcolsep}{4pt}
\begin{tabular}{l cc ccc}
\toprule
& \multicolumn{2}{c}{\textbf{Epoch range}} & \multicolumn{3}{c}{\textbf{Ramped weights} [start $\to$ end]} \\
\cmidrule(lr){2-3} \cmidrule(lr){4-6}
\textbf{Task} & Start & End & $w_{\text{collision}}$ & $w_{\text{near\_cent.}}$ & $w_{\text{centering}}$ \\
\midrule
State-Static  & 500  & 2500 & $0 \to 18$  & $0 \to 3$  & $3 \to 3$ \\
State-Moving  & 2000 & 6000 & $0 \to 70$  & $0 \to 25$ & $3 \to 6$ \\
\bottomrule
\end{tabular}
\caption{PPO reward curriculum schedule. Weights are linearly annealed
over the specified epoch range.}
\label{tab:ppo_curriculum}
\end{table}

\end{document}